\documentclass[10pt,twocolumn,letterpaper]{article}

\usepackage{iccv}
\usepackage{times}
\usepackage{epsfig}
\usepackage{graphicx}
\usepackage{amsmath}
\usepackage{amssymb}

\usepackage{epsfig}
\usepackage{graphicx}
\usepackage{comment}

\usepackage{floatrow}
\newfloatcommand{capbtabbox}{table}[][\FBwidth]

\usepackage{graphicx}
\makeatletter
\newcommand{\shorteq}{%
  \settowidth{\@tempdima}{-}
  \resizebox{\@tempdima}{\height}{=}%
}

\usepackage[pagebackref=true,breaklinks=true,letterpaper=true,colorlinks,bookmarks=false]{hyperref}

\iccvfinalcopy 

\def\iccvPaperID{2993} 

\ificcvfinal\fi
\begin{document}

\title{Explicit Spatiotemporal Joint Relation Learning for Tracking Human Pose}

\author{Xiao Sun\\
Microsoft Research\\
{\tt\small xias@microsoft.com}
\and
Chuankang Li \thanks{Work done during internship at Microsoft Research Asia}\\
Zhejiang University\\
{\tt\small lck-cad@zju.edu.cn}
\and
Stephen Lin\\
Microsoft Research\\
{\tt\small stevelin@microsoft.com}
}

\maketitle

\begin{abstract}
We present a method for human pose tracking that is based on learning spatiotemporal relationships among joints. Beyond generating the heatmap of a joint in a given frame, our system also learns to predict the offset of the joint from a neighboring joint in the frame. Additionally, it is trained to predict the displacement of the joint from its position in the previous frame, in a manner that can account for possibly changing joint appearance, unlike optical flow. These relational cues in the spatial domain and temporal domain are inferred in a robust manner by attending only to relevant areas in the video frames. By explicitly learning and exploiting these joint relationships, our system achieves state-of-the-art performance on standard benchmarks for various pose tracking tasks including 3D body pose tracking in RGB video, 3D hand pose tracking in depth sequences, and 3D hand gesture tracking in RGB video.
\end{abstract}

\section{Introduction}
\label{sec.intro}

Human pose tracking is an essential component of many video understanding tasks such as visual surveillance, action recognition, and human-computer interaction. Compared to human pose estimation in single images, pose tracking in videos has shown to be a more challenging problem due in part to issues such as motion blur and uncommon poses with few or no examples in training sets. To help alleviate these problems, temporal information and spatial structure have been utilized in locating body joints.

A variety of approaches have been presented for exploiting temporal information in tracking human pose. Among them are techniques that use dense optical flow to propagate joint estimates from previous frames~\cite{jain2014modeep,pfister2015flowing,song2017thin}, 
local model fitting with the predicted pose of the preceding frame as initialization~\cite{dabral2018learning,bogo2016keep,mehta2017vnect,qian2014realtime,oikonomidis2011efficient}, and temporal smoothness priors as a pose estimation constraint~\cite{zhou2016sparseness,mueller2018ganerated}. 
These methods take advantage of pose similarity from frame to frame, but do not learn explicitly about the effect of human motions on the appearance and displacement of joints. Rather, the motions are predicted through generic machinery like optical flow, local optimization of an initialized model, and smoothness constraints. 

Other methods take advantage of the spatial structure of joints to facilitate joint localization in tracking. These include techniques that integrate graphical models into deep networks~\cite{tompson2014joint, ouyang2013joint, yang2016end} and that employ structural priors~\cite{zhou2017towards, dabral2018learning}. Such approaches are mainly designed to avoid invalid pose configurations rather than directly localize joints. Alternatively, several methods refine part locations based on the predicted positions of other parts~\cite{bulat2016human, wei2016convolutional, newell2016stacked, gkioxari2016chained}, which provides only vague constraints on joint positions.

In this paper, we propose an approach that \emph{explicitly} learns about spatiotemporal joint relationships and utilizes them in conjunction with conventional joint heatmaps to elevate performance on human pose tracking.
For learning temporal relationships, our system design reconfigures the optical flow framework to introduce elements from human pose estimation. Instead of densely predicting the motion vectors of general scene points as done for optical flow, our network is trained specifically to infer the displacements only of human joints, while also accounting for possible appearance changes due to motion. 
Inspired by heat maps in pose estimation, we leverage the other pixels to support the inference of each joint displacement. With training on labeled pose tracking datasets, our neural network learns to make use of information from both joint and non-joint pixels, as well as appearance change caused by joint motion, to predict joint displacements from one frame to the next. 

Our system additionally learns spatial joint relationships, specifically the offset between a joint and a particular neighboring joint. These spatial relationships are learned in a manner analogous to the temporal relationships, except in the spatial domain. The spatial and temporal relationships, as well as single-frame joint heatmaps, all provide orthogonal cues based on different source information for predicting joint locations. We show that optimizing with respect to these cues together leads to appreciable improvements in pose tracking performance. 

The presented technique comes with the following practical benefits:
\begin{itemize}
\item {\it Generality:} This approach can be employed for various pose tracking applications (body/hand, 2D/3D, with/without an object) and with different input modalities (RGB/depth). In our experiments, the system is demonstrated on 3D human body pose tracking in RGB video, 3D hand gesture tracking from depth sequences, and 3D hand pose tracking in RGB video.
\item {\it Flexibility:} The learned spatiotemporal relations are complementary not only to heatmap-based pose estimation, but also to other approaches such as local model fitting. Its speed-accuracy tradeoff is also controllable.
\item {\it High performance:} State-of-the-art results are obtained on standard benchmarks for the aforementioned pose tracking applications, as shown in our experiments.
\end{itemize}
The code and model will be released online upon publication of this work.


\section{Related Work}
\label{sec.rela}

Related research has exploited \emph{joint structure} for single-frame human pose estimation and \emph{temporal information} for pose tracking in video. We review these techniques, as well as methods for joint association within a multi-person context.

\textbf{Joint Structure.}
Several methods incorporate structural constraints on joint locations by integrating graphical models~\cite{felzenszwalb2005pictorial} into deep networks. These graphical models have been used to remove anatomically incorrect outliers from part detections~\cite{tompson2014joint}, constrain articulation with deformation models~\cite{ouyang2013joint}, 
and learn the spatial compatibility of neighboring parts to regularize the output~\cite{yang2016end}. The graphical constraints in these works are employed to enforce global pose consistency, rather than to provide explicit estimates of relative part positions.


In some other works, estimates of part locations are refined based on intermediate predictions of other parts. This refinement has been done by regressing part locations from the full set of part detection heatmaps~\cite{bulat2016human}, by iteratively computing the set of heatmaps within a multi-stage architecture~\cite{wei2016convolutional,newell2016stacked}, and by sequentially predicting joint heatmaps using previously predicted parts in order of difficulty~\cite{gkioxari2016chained}. Refinement has also been performed in the feature maps of different joints via message passing~\cite{chu2016structured}. In these methods, the intermediate part predictions provide only implicit constraints on joint localization.


Methods have also re-parameterized the pose representation to model joint structure. Such representations have been based on an overcomplete 3D pose dictionary~\cite{tekin2016structured} and kinematic models of the human body~\cite{zhou2016deep} and hand~\cite{zhou2016model}. These re-parameterizations are either high-dimensional or highly nonlinear, which makes optimization in deep networks hard. Alternatively, a geometric loss function has been introduced to enforce consistency of bone-length ratios in the human body~\cite{zhou2017towards}, and this has been extended to include joint-angle limits and left-right symmetry~\cite{dabral2018learning}. Similar to the constraints of graphical models, these structural priors mainly serve to restrict pose predictions to valid configurations.


Recently, a practical pose re-parameterization based on bones instead of joints has been proposed together with a compositional loss~\cite{sun2017compositional}. We utilize these simple bone vectors to represent joint relationships in the spatial domain, and explicitly use bones as one of our learning targets. In~\cite{sun2017compositional}, the performance is limited by its use of fully connected regression that maps the entire image feature to a local bone prediction. For better estimation, we instead propose to use a per-pixel regression together with an attention-like pixel weighting mechanism so as to take full advantage of the most relevant pixels while ignoring irrelevant ones. 

\textbf{Temporal Information.} Many deep learning methods enhance pose estimation in the current frame with the help of dense optical flow. Among them are methods that concatenate the flow field with RGB features~\cite{jain2014modeep}, compute flow-based alignment of heatmaps from previous frames to the current frame~\cite{pfister2015flowing}, and define temporal graph edges using flow~\cite{song2017thin}. Reliance on dense optical flow can be problematic, as flow estimates may be unreliable for non-rigid body regions that undergo appearance change during motion. In our work, we specifically train our model to predict the displacement only of human joints, with the support of information from relevant local areas. This task-specific approach provides greater reliability than generic flow, as shown empirically in Section~\ref{fig:human3d}. In addition, this approach is directly applicable to both 2D and 3D pose, while optical flow provides only 2D cues.

Temporal information is also exploited in techniques built on recurrent neural networks (RNNs)~\cite{lin2017recurrent, coskun2017long, hossain2017exploiting, luo2017lstm}, which account for information from previously predicted poses in determining the solution for the current frame. Similarly, some works fully connect the input and output pose sequences directly using a fully connected~\cite{dabral2018learning,einfalt2018activity} or convolutional~\cite{gkioxari2016chained} network, which draws global dependencies between the input and output. In contrast to these methods which implicitly encode temporal dependency into a network, we explicitly regress temporal joint displacements from only the most relevant input pixels. Attending only to information relevant to what is being processed has shown greater efficacy than recurrent computation in certain tasks~\cite{vaswani2017attention}.

Some works follow the \emph{tracking by detection} paradigm, which performs detection in each frame, estimates the pose of each detection result with an off-the-shelf technique, then enforces spatio-temporal smoothness for tracking~\cite{zhou2016sparseness, mueller2018ganerated}. Instead of employing generic smoothness priors, we seek a more targeted approach by specifically learning about human pose change.


\textbf{Multi-Person Joint Association.}
Relationships between joints are also considered by methods for multi-person tracking and grouping. For the multi-person tracking problem, the goal is to estimate the pose of all persons appearing in the video and assign a unique identity to each person~\cite{iqbal2016posetrack,girdhar2018detect,xiao2018simple,xiu2018pose,fabbri2018learning,insafutdinov2017arttrack,doering2018joint}. These approaches differ from one another in their choice of metrics and features used for measuring similarity between a pair of poses and in their algorithms for joint identity assignment. We note that in these techniques, the concept of `tracking' is to associate joints of the same person together over time, using joints localized independently in each frame. By contrast, our work aims to improve joint localization by utilizing information from other frames.


Multi-person grouping aims to link the body parts of the same person in the context of multi-person pose estimation. To associate joints within an image, part affinity fields~\cite{cao2016realtime} and associative embedding tags~\cite{newell2017associative} have been learned together with detection heatmaps. This information is then used to group body parts into separate identities. While these techniques learn joint relationships in terms of their association, our work instead focuses on explicitly predicting offsets between neighboring joints.

\section{Explicit Spatiotemporal Joint Relation Learning}
\label{sec.cond}



Given a video sequence, the pose tracking problem is to obtain the 2D (or 3D) position of all the $K$ joints in all $T$ frames, $\mathcal{J}=\{\mathbf{J}_k^t|k=1,...,K; t=1,...,T\}$. 
In solving this problem, we seek to supplement the per-joint heatmaps conventionally predicted from a single frame with {\bf explicit} predictions of spatiotemporal joint relations. These relations include both the offset of a joint from a neighboring joint in the same frame and the offset of the joint from its position in the previous frame.



\begin{figure}
\includegraphics [width=1.1\linewidth] {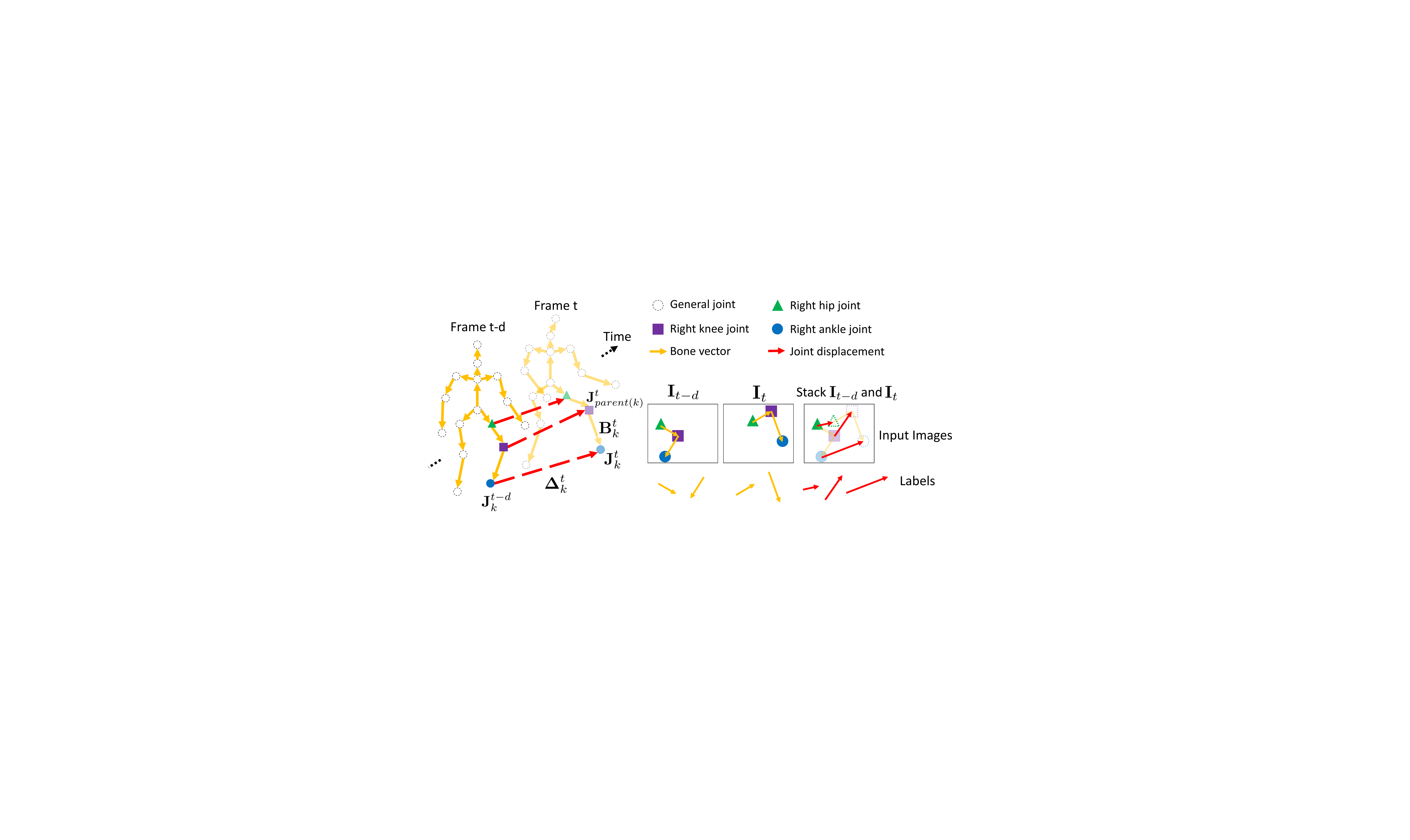}
\caption{Spatiotemporal joint relation learning problem.}%
\label{fig.definition}
\end{figure}

In modeling joint relations, we utilize a relation function $relation(k,t)$ that returns the index of the related joint for the $k^{th}$ joint at time $t$, in either the spatial or temporal domain. This function is defined as
\begin{equation}
\label{eq.related_joint}
relation(k,t) = \left\{ \begin{array}{rl}
(parent(k), \;\;\;t\;\;\;), &\mbox{in spatial domain} \\
(\;\;\;\;\;\;\;k\;\;\;\;\;\;, t-d), &\mbox{in temporal domain}
\end{array} \right.
\end{equation}
where $parent(k)$ returns the index of the parent joint for the $k^{th}$ joint in a joint tree illustrated in Figure~\ref{fig.definition} (left).

The relationship between joint $\mathbf{J}_k^t$ and its related joint $\mathbf{J}_{relation(k,t)}$ is defined as a directed vector pointing from the related joint to it:
\begin{equation}
\label{eq.relation}
\mathbf{R}^{t}_k = \mathbf{J}^{t}_k - \mathbf{J}_{relation(k,t)}.
\end{equation}
Combining Eq.~\ref{eq.relation} and Eq.~\ref{eq.related_joint}, we have
\begin{equation}
\label{eq.relation_joint}
\mathbf{R}^{t}_k = \left\{ \begin{array}{rl}
\mathbf{B}^t_k = \mathbf{J}^t_k - \mathbf{J}^t_{parent(k)}, &\mbox{for spatial relation} \\
\mathbf{\Delta}^{t}_k = \mathbf{J}^{t}_k - \mathbf{J}^{t-d}_k\;\;\;\;\;\;\;\,, &\mbox{for temporal relation.}
\end{array} \right.
\end{equation}
The spatial joint relation $\mathbf{B}^t_k$ is equivalent to the \emph{\textbf{bone vector}} in Sun et al.~\cite{sun2017compositional}, where the associated bone of the $k^{th}$ joint is defined as the vector pointing from its parent to it. The temporal joint relation $\mathbf{\Delta}^{t}_k$ represents the \emph{\textbf{joint displacement}} from one time instance ($t-d$) to the next ($t$). 

Given two successive frames $\mathbf{I}_{t-d}$ and $\mathbf{I}_{t}$ from a video sequence, our goal is to explicitly estimate the bone vector and joint displacement for each joint as shown in Figure~\ref{fig.definition} (right). Note that joint displacement involves the frames at both $t$ and $t-d$. We thus stack $\mathbf{I}_{t-d}$ and $\mathbf{I}_{t}$ as input. On the other hand, the bone vector is computed from only a single input frame ($\mathbf{I}_{t-d}$ or $\mathbf{I}_{t}$). The notations used in our spatiotemporal joint relation learning are summarized in Table~\ref{table.notation}.

\begin{table}
\caption{Notations.}
\begin{center}
\footnotesize
\begin{tabular}{l | l | l | l | l }
\hline
         & Joint & Related joint &  Learning target: Relation & Input\\
\hline
General & $\mathbf{J}_{k}^t$ & $\mathbf{J}_{relation(k,t)}$ & $\mathbf{R}^{t}_k~\shorteq~\mathbf{J}^{t}_k$ - $\mathbf{J}_{relation(k,t)}$ & - \\
\hline
Spatial  & $\mathbf{J}_{k}^t$ & $\mathbf{J}_{parent(k)}^t$ & $\mathbf{B}^t_k~\shorteq~\mathbf{J}^t_k$ - $\mathbf{J}^t_{parent(k)}$ &  $\mathbf{I}_{t}$ \\
\hline
Temporal & $\mathbf{J}_{k}^t$ & $\mathbf{J}_{k}^{t-d}$ & $\mathbf{\Delta}^{t}_k~\shorteq~\mathbf{J}^{t}_k$ - $\mathbf{J}^{t-d}_k$ & $\mathbf{I}_{t-d}$,$\mathbf{I}_{t}$ \\
\hline
\end{tabular}
\end{center}
\label{table.notation}
\end{table}

 





\subsection{Fully Connected Regression: A Baseline}
A possible way to predict a relation $\mathbf{R}_k^t$ is by using a CNN with a fully connected layer as done in~\cite{sun2017compositional} for predicting bone vectors. The input images are passed through convolutional layers to generate convolutional features. Following common practice~\cite{sun2017compositional,carreira2016human,zhou2016deep,zhou2017towards,sun2018integral}, the features undergo spatial downsampling by average pooling. They are then given to a \emph{\textbf{fully connected}} layer which outputs a $3K$-dimensional bone or joint displacement vector. We refer to this holistic fully connected approach as the \textbf{\emph{Baseline}}.

There is a clear drawback in the \textbf{\emph{Baseline}} approach. Image features and outputs are fully connected, but most image areas/features are \emph{irrelevant} to a specific relation. For example, pixels that lie far from the joint's vicinity are generally less informative for this prediction. 
The inclusion of non-relevant pixels and the over-emphasis of weakly relevant pixels would not only be unhelpful but could moreover be detrimental due to the noise they would introduce to the estimation. Worse still, the relevant image areas can change rapidly with respect to different joint locations, which would make the learning even more difficult.








\subsection{Per-pixel Regression with Attention}
\label{sec.representation}

To address the above issues, we present a method for per-pixel joint relation regression with an attention-like pixel weighting mechanism.



\begin{figure}
\includegraphics [width=1.1\linewidth] {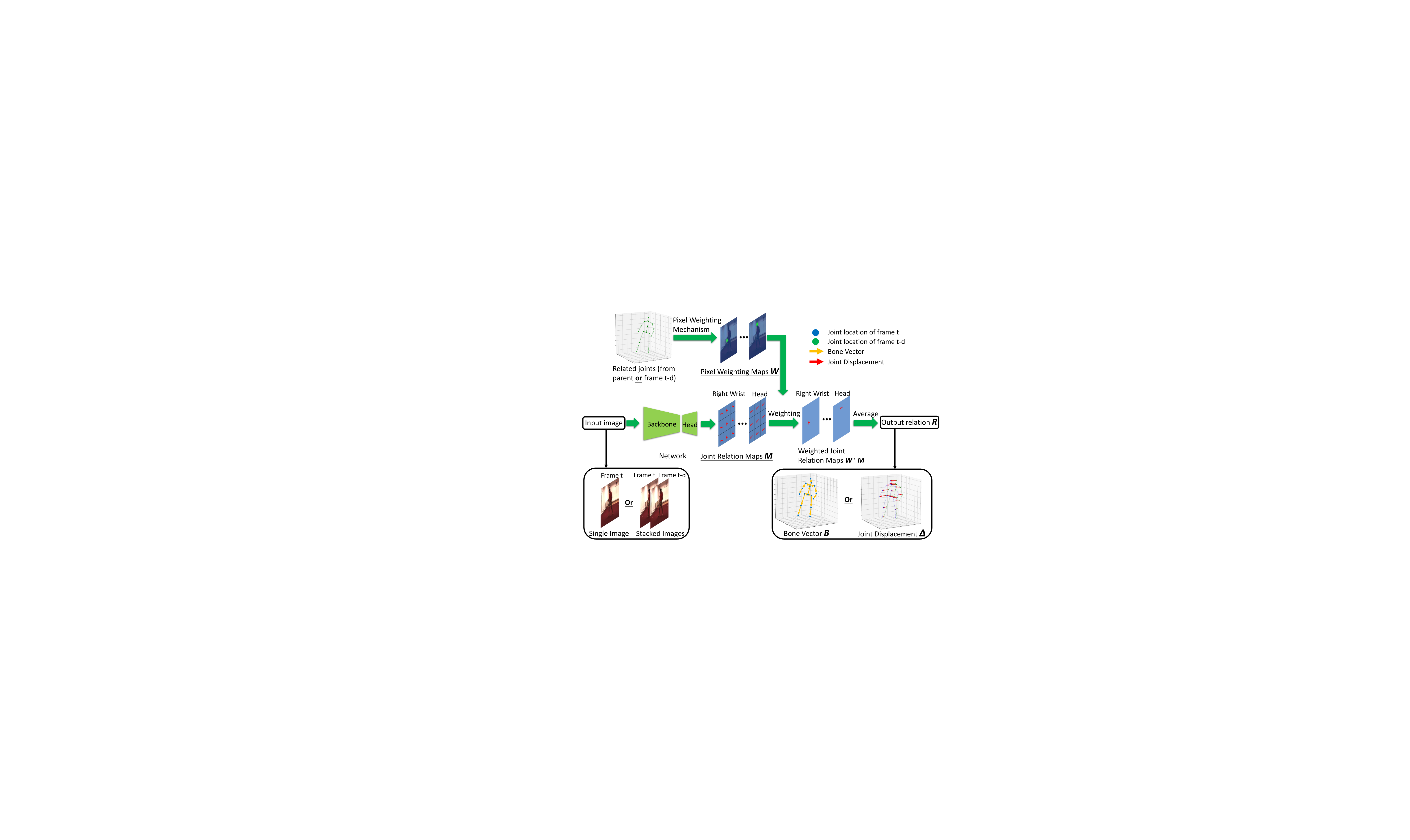}
\caption{Overview of the explicit joint relation learning network.}%
\label{fig.overview}
\end{figure}

\textbf{Joint Relation Map.} Similar to heatmap based pose estimation, we generate a \emph{\textbf{separate}} relation map $\mathbf{M}_k^t$ for each joint, where every pixel in a map predicts the relation $\mathbf{R}_k^t$ of the joint, as illustrated in Figure~\ref{fig.overview}. For 3D pose, a joint relation map is a 3D vector field for the x, y, z dimensions. For compactness, the joint index $k$ and time index $t$ will be omitted in the rest of this section, as all joints at all times are processed separately and in the same way.

\textbf{Pixel Weighting Map.} However, not every pixel in a joint relation map is useful for estimating the relation of the joint. Therefore, an attention mechanism for filtering out these less informative pixel predictions is needed. To represent this information, a pixel weighting map $\mathbf{W}$ is generated for each joint relation map, as shown in Figure~\ref{fig.overview}. Each pixel in $\mathbf{W}$ contains a weight value in $[0, 1]$ that indicates the relevance of the pixel for predicting the relation of this joint. 

\begin{figure}
\includegraphics [width=0.98\linewidth] {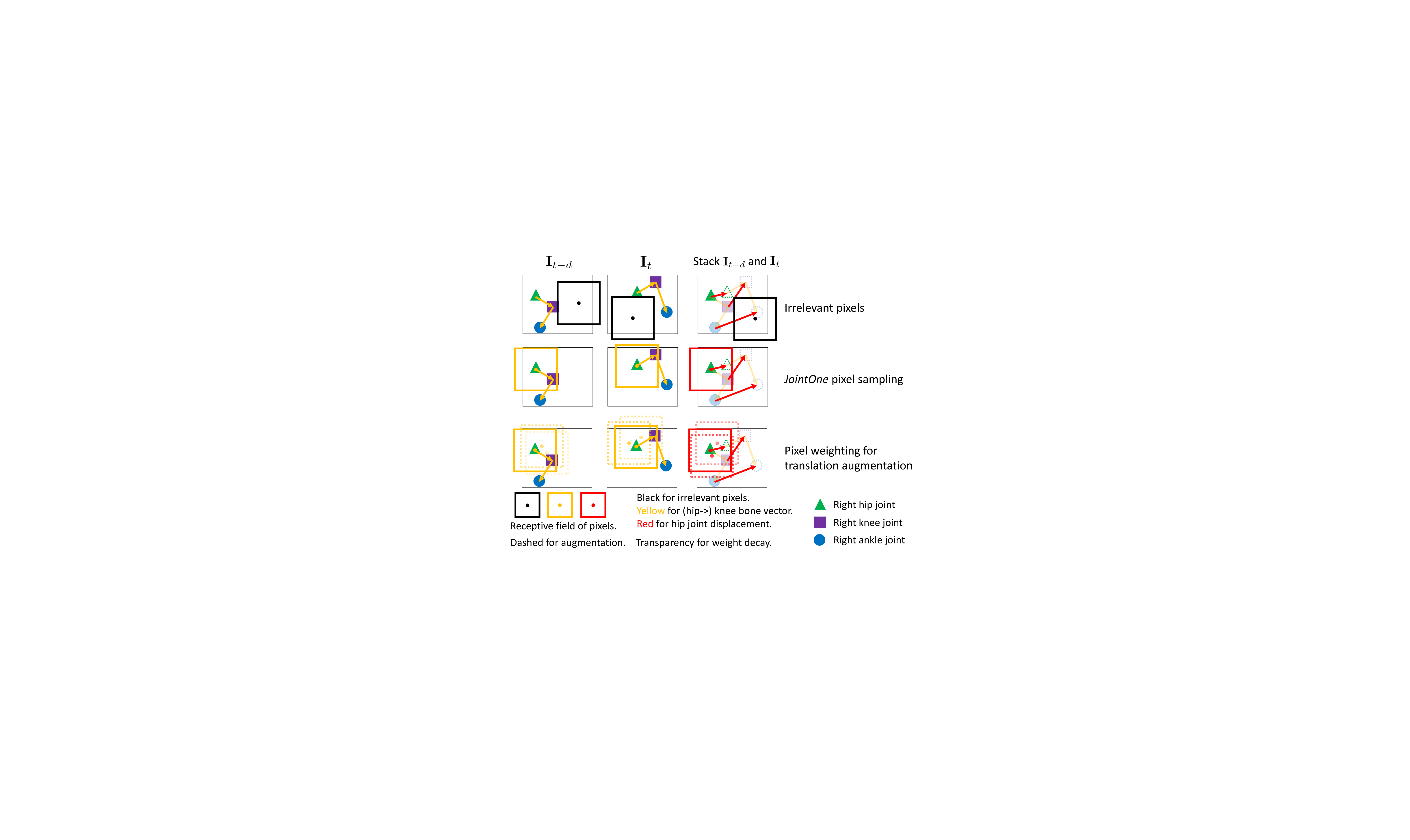}
\caption{Per-pixel regression on the most relevant pixels.}%
\label{fig.pixel_weighting}
\end{figure}

Intuitively, pixels closer to the related joint are generally more informative for estimating the joint relation. We denote \emph{JointOne} as taking only one pixel at the ground truth location of the related joint $\mathbf{J}_{relation(k,t)}$. For example, Figure~\ref{fig.pixel_weighting} (middle row) shows the \emph{JointOne} pixel sampling for (hip$\rightarrow$) knee bone vector (yellow box) and hip joint displacement (red box) regression.

At the testing phase, the ground truth location of $\mathbf{J}_{relation(k,t)}$ is not available. Instead, we use its prediction from the heat map estimation. To make the relation learning more robust to inaccurate heat map prediction, we propose a general pixel weighting formulation which is similar to pixel \textbf{\emph{translation augmentation}} as shown in Figure~\ref{fig.pixel_weighting} (bottom row), where box transparency indicates the weight decay of a pixel sample. Formally,
\begin{equation}
\label{eq.weight_map_general}
\mathbf{W} = \mathcal{O}(\mathbf{F}, \beta)
\end{equation}
where $\mathcal{O}$ is a decay function that determines the drop in weight for pixels farther away from $\mathbf{J}_{relation(k,t)}$. $\mathbf{F}$ is a distance map in which each pixel value represents the distance from the pixel to $\mathbf{J}_{relation(k,t)}$. $\beta$ is the decay rate parameter. For the case of an exponential decay function, the pixel weighting map becomes 
\begin{equation}
\label{eq.exp_decay}
\mathbf{W} = \mathcal{O}(\mathbf{F}, \beta) = e^{-\beta \mathbf{F}}.
\end{equation}
Several decay functions are empirically investigated in Section~\ref{sec.exp}.

\textbf{Discussion.} For a fully convolutional network, per-pixel regression essentially generates local image patches over each of the pixels using a sliding window and regresses the relation $\mathbf{R}$ from each local patch. The size of the sliding window is equal to the receptive field of the network. As shown in Figure~\ref{fig.pixel_weighting}, the per-pixel regression formulation combined with the pixel weighting mechanism is an effective way to filter out irrelevant pixel/patch samples and only preserve informative ones.

\begin{figure}
\includegraphics [width=1.05\linewidth] {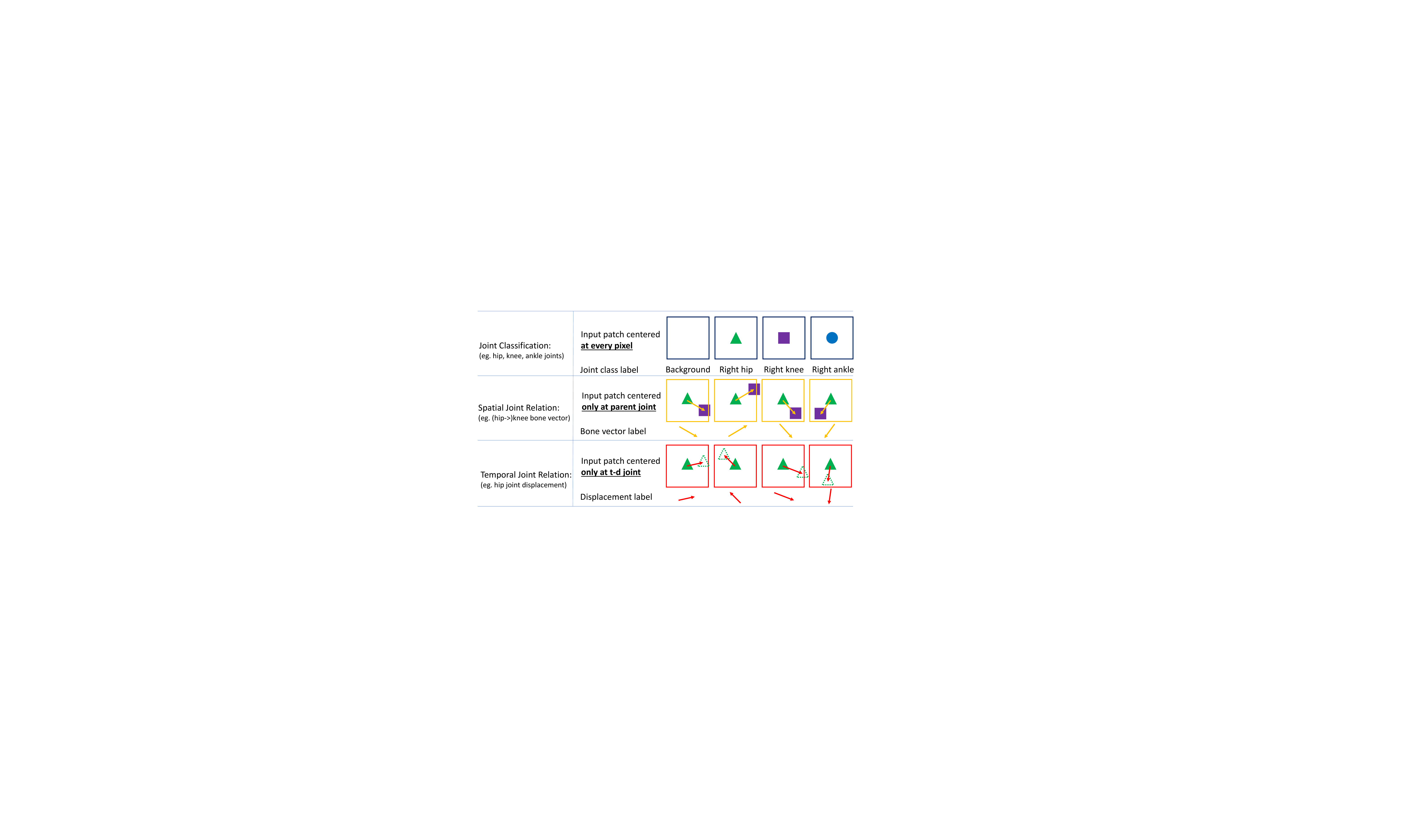}
\caption{Orthogonal perspectives of joint classification, spatial joint relation regression, and temporal joint relation regression.}%
\label{fig.task_relationship}
\end{figure}

Figure~\ref{fig.task_relationship} illustrates on a local patch the difference in information provided by heatmap based per-pixel joint classification and our explicit spatiotemporal joint relation regression. It can be seen that these three tasks approach the joint localization problem from orthogonal perspectives and using different observational data:
\begin{itemize}
    \item \textbf{Joint Classification.} Determining whether a pixel is the correct type of joint from its local image appearance.
    \item \textbf{Spatial Joint Relation Regression.} Predicting a child joint position \textbf{\emph{conditioned}} on its parent joint location.
    \item \textbf{Temporal Joint Relation Regression.} Estimating the change in joint location based on the \textbf{\emph{temporal transition}}.
\end{itemize}

Our spatiotemporal relation learning effectively makes use of additional joint localization cues (dependence in space and continuity in time). The explicitly predicted bone vectors and joint displacements are thus \emph{\textbf{complementary}} to single-frame per-joint detection results which are determined from only static joint appearance. Through integrating these cues, there exists a stronger foundation for inferring joint locations, via an optimization presented in Section~\ref{sec.tracking}.


\textbf{Sparse Flow Baseline.} Our explicit learning of temporal relationships is related to optical flow estimation, but it alters the conventional optical flow scheme to incorporate concepts from pose estimation. Rather than computing the motion of each pixel to form one dense flow map, our system focuses on predicting the motion only of pixels with specific semantics, namely joints. What's more, a \emph{\textbf{separate}} joint displacement map is generated for each joint, containing displacements predicted for that joint only. For an expanded comparison to optical flow, we introduce another baseline, called \emph{\textbf{SparseFlow}}, that lies between optical flow and our method. It generates a \textbf{\emph{single}} flow map as in optical flow but only predicts the motion of pixels at joint locations as in our method. A comprehensive comparison between optical flow, \emph{\textbf{SparseFlow}} and our method is shown in Section~\ref{sec.exp}.

\subsection{Unified Network Architecture}
\label{sec.architecture}
We use a unified network architecture for both spatial and temporal joint relation learning as shown in Figure~\ref{fig.overview}. The only difference lies in the input. Spatial joint relation uses a single image as input while temporal joint relation stacks two consecutive images together as input. The input images are fed into a fully convolutional network. Then, a head network consisting of consecutive deconvolution layers is used to upsample the feature map to the targeted resolution as done in~\cite{sun2018integral}. At the end, a 1x1 convolutional layer is used to produce the final joint relation maps.

\textbf{Training Loss.}  We directly minimize the absolute difference (L1 norm) of the predicted and ground truth joint relation maps weighted by the pixel weighting map. This loss is expressed as
\begin{equation}
\label{eq.loss_func}
loss = \int_{\mathbf{p}\in\Omega}(\mathbf{W} \cdot ||\mathbf{M}^{pre} - \mathbf{M}^{gt}||_1)
\end{equation}
where $\Omega$ is the domain of the joint relation map. For other training details, please see the supplement.

\textbf{Inference} In the testing phase, the final prediction is the average of all predictions in a joint relation map weighted by the pixel weighting map:
\begin{equation}
\label{eq.inference}
\mathbf{R} = \dfrac{\int_{\mathbf{p}\in\Omega}(\mathbf{W} \cdot \mathbf{M}^{pre})}{\int_{\mathbf{p}\in\Omega}\mathbf{W}}.
\end{equation}

\section{Spatiotemporal Joint Relations for Tracking}
\label{sec.tracking}


Given single-frame per-joint pose predictions $\mathbf{J}_k^t$ and spatiotemporal joint relation predictions $\mathbf{B}_k^t$ and $\mathbf{\Delta}_k^t$, the pose tracking task can be formulated as a simple linear optimization problem. For a video sequence with $T$ frames, we minimize the following error function to obtain the final tracking result $\mathbf{\hat{J}}_k^t$:


\begin{equation}
\label{eq.err_func_pro}
\begin{split}
e = \sum_{t=1}^{T} \sum_{k=1}^{K} (||\mathbf{\hat{J}}_k^t - \mathbf{J}_k^t||^2 & + \alpha * ||\mathbf{\hat{J}}_k^t - \mathbf{\hat{J}}_{parent(k)}^t - \mathbf{B}_k^t||^2 \\
 & + \sum_{n=1}^{N} (\gamma_n * ||\mathbf{\hat{J}}_k^t - \mathbf{\hat{J}}_k^{t-\mathbb{D}(n)} - \mathbf{\Delta}_k^t||^2)) \\
\end{split}
\end{equation}
where $\alpha$ and $\gamma$ are parameters to balance the single-frame and joint relation terms. 



This optimization can be applied not only to two consecutive frames, but also to frames separated by different durations. Accounting for additional frames can potentially enhance the pose tracking. For this purpose, we define a set $\mathbb{D}$ which represents $N$ timesteps of duration $d$. 

Minimizing $e$ with respect to $\mathbf{\hat{J}}_k^t$ is a linear least squares problem with $(N+2)*K*T$ constraints, which is sufficient to solve the $K*T$ unknowns. The duration set $\mathbb{D}$ can be arbitrary. Experimentally, we will consider four variants:

\begin{itemize}
\item $\mathbb{D}_{f} = \{1\}$, which is the common case of considering temporal joint relations only from the previous frame to the current frame (corresponding to one-step forward tracking).
\item $\mathbb{D}_{fb} = \{1,-1\}$, which additionally considers displacements from the subsequent frame to the current frame (one-step forward and backward tracking).
\item $\mathbb{D}_{mf} = \{1,2,3\}$, which accounts for temporal joint displacements from multiple past frames (multi-step forward tracking).
\item $\mathbb{D}_{mfb} = \{1,2,3,-1,-2,-3\}$, which utilizes displacements from multiple past and future frames (multi-step forward and backward tracking).
\end{itemize}

\section{Experiments}
\label{sec.exp}
Our approach can be applied to various pose tracking scenarios, including human body or hand (w/ or w/o an object), RGB or depth input, and 2D or 3D output. We evaluate our approach extensively on four challenging datasets that cover all of these factors to show its generality. Specifically, 3D human pose dataset Human3.6M~\cite{ionescu2014human3} of RGB images and 3D hand pose dataset MSRA Hand 2015~\cite{sun2015cascaded} of depth images are used in the main paper. 3D hand pose datasets Dexter+Object~\cite{sridhar2016real} and Stereo Hand Pose~\cite{zhang20163d} of RGB-D images are used in the supplement.

\subsection{Experiments on Human3.6M}
\label{fig:human3d}


Human3.6M~\cite{ionescu2014human3} (HM36) is the largest 3D human pose benchmark of RGB images. It consists of 3.6 million video frames. 
For this benchmark, we employ the most widely used evaluation protocol in the literature~\cite{chen20163d,tome2017lifting,moreno20163d,zhou2017monocap,jahangiri2017generating,mehta2016monocular,pavlakos2016coarse,yasin2016dual,rogez2016mocap,bogo2016keep,zhou2016sparseness,tekin2016direct,zhou2016deep,sun2018integral,sarandi2018synthetic}. Five subjects (S1, S5, S6, S7, S8, S9) are used in training and two subjects (S9, S11) are used in evaluation. The video sequences are typically downsampled for efficiency. We use three downsampling rates, namely no downsampling (25FPS), three-step downsampling (8FPS) and ten-step downsampling (2.5FPS). The lower the FPS, the larger the joint displacement between two consecutive frames.


For evaluation, most previous works use the mean per joint position error (\emph{MPJPE}) in millimeters (mm). We call this metric \emph{\textbf{Joint Error}}. To directly evaluate spatiotemporal joint relation performance, we use two additional metrics. The first metric is the mean per-bone position error, or \emph{\textbf{Bone Error}} as in~\cite{sun2017compositional}. The second metric is the mean per-joint displacement error, or \emph{\textbf{Displ Error}}, which adds the estimated joint displacement to the ground truth pose of the previous frame to obtain the current frame pose estimation and then evaluate its \emph{Joint Error} in mm.


\textbf{Ablation Study} We first determine the best settings of our model for the explicit learning of spatiotemporal joint relations. For this purpose, we only use the joint relation evaluation metrics, namely \emph{Bone Error} and \emph{Displ Error}, because they isolate joint relation estimation from other factors such as the quality of single frame pose estimation and the specific tracking algorithm used. 

\begin{figure}
\ffigbox[1.0\textwidth]
{
\includegraphics [width=0.494\linewidth] {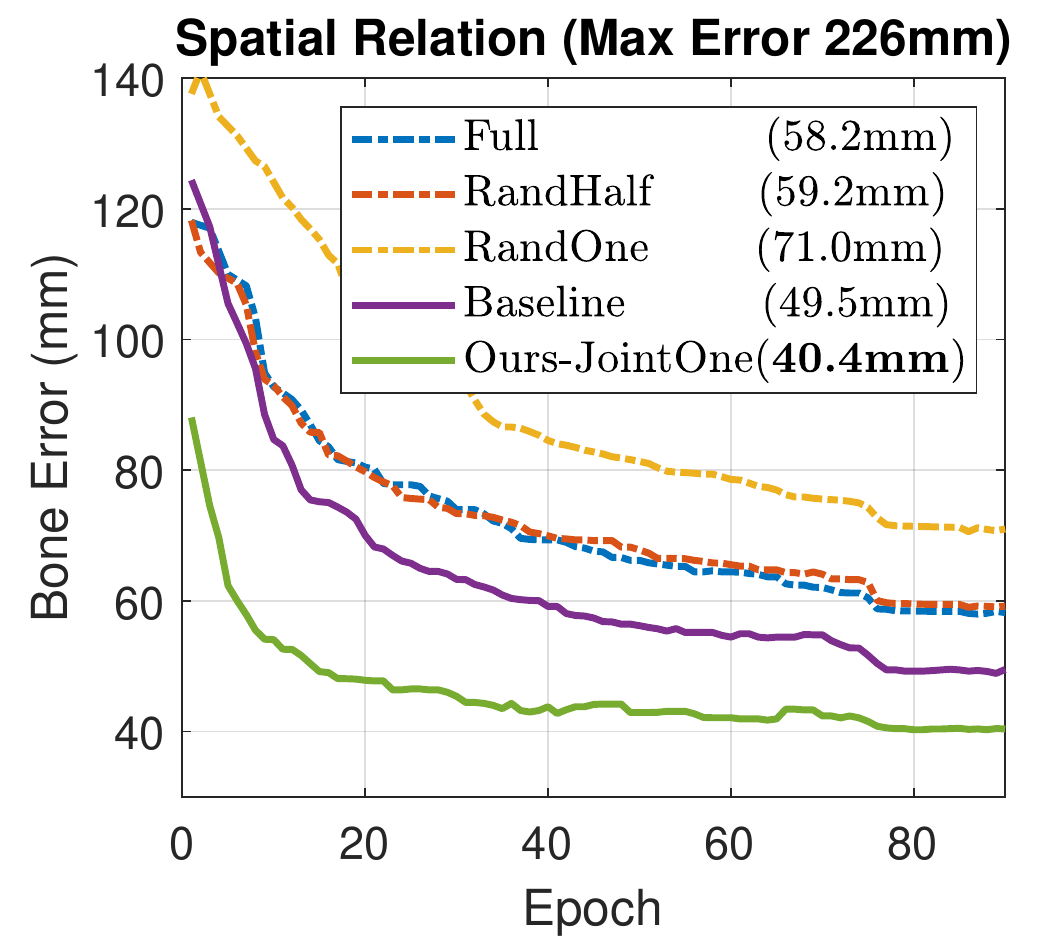}
\includegraphics [width=0.494\linewidth] {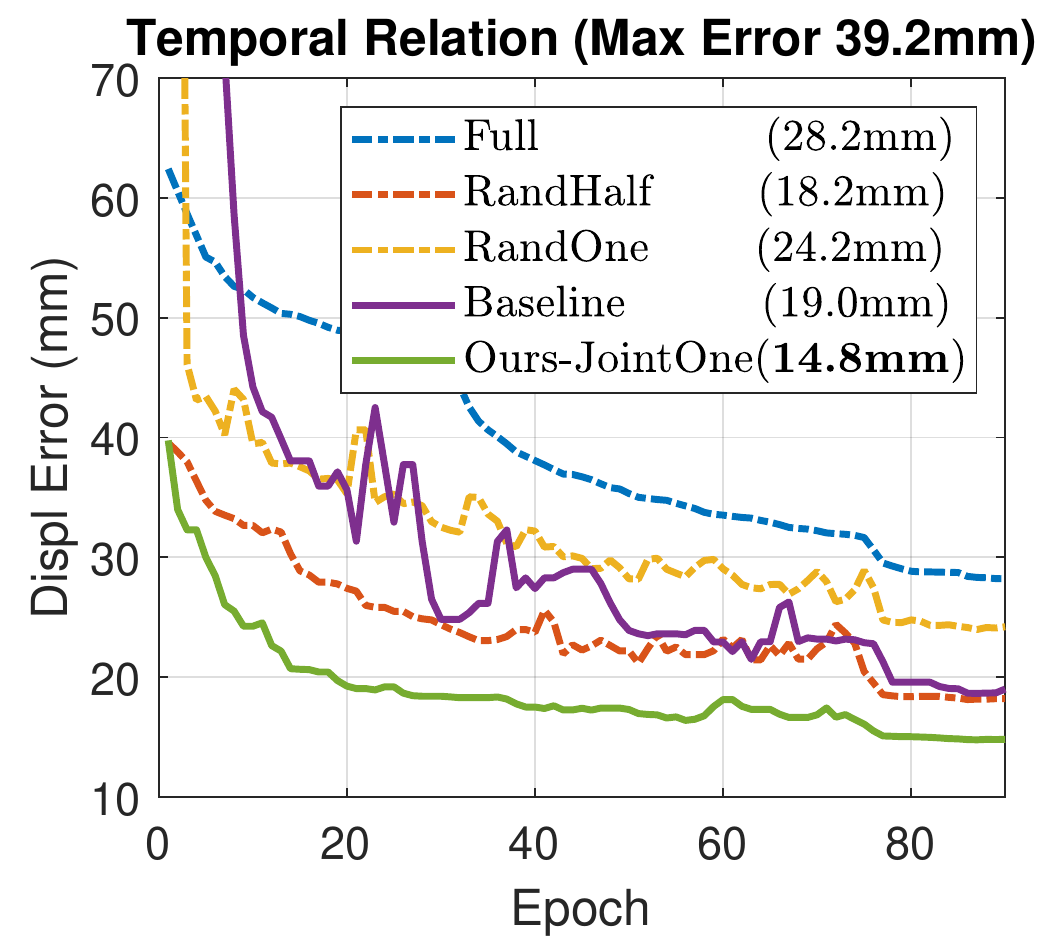}
}
\quad
\capbtabbox[1.05\textwidth]
{
\footnotesize
\begin{tabular}{l | l | l | l | l | l }
\hline
Error (mm) & Full & RandHalf & RandOne & Baseline & Ours-JointOne \\
\hline
\textbf{Bone Error}& 58.2 & 59.2 & 71.0 & \underline{49.5} & $\textbf{40.4}_{\downarrow18.4\%}$ \\
\hline
\textbf{Displ Error}& 28.2 & 18.2 & 24.2& \underline{19.0} & $\textbf{14.8}_{\downarrow22.1\%}$    \\
\hline
\end{tabular}
}
{
  \caption{Comparison of the fully connected \emph{\textbf{Baseline}} and our per-pixel regression. Different pixel sampling strategies are investigated.}
  \label{table.pixel_sampling}
}
\end{figure}
\begin{figure}
\begin{floatrow}
\capbtabbox[.45\textwidth]
{%
  \small
  \begin{tabular}{c|c}
  \hline
  function & formulation \\ 
  \hline
   Binary & $ \begin{cases} 1 & \text{if } \mathbf{J} \leq \beta \\ 0 & \text{if } \mathbf{J} > \beta \end{cases} $\\
  \hline
  Gaussian & $ e^{-\beta \mathbf{J}^2} $ \\
  \hline
  Linear & $ 1 - \beta \mathbf{J} $\\
  \hline
  Exponential & $ e^{-\beta \mathbf{J}} $ \\
  \hline
  \end{tabular}
}{%
  \caption{Different definitions of the decay function $\mathcal{O}(\mathbf{J}, \beta)$ for Eq.~\ref{eq.weight_map_general}.}%
  \label{table.decay_func}
}
\ffigbox[.45\textwidth]
{%
  \includegraphics [width=1.1\linewidth] {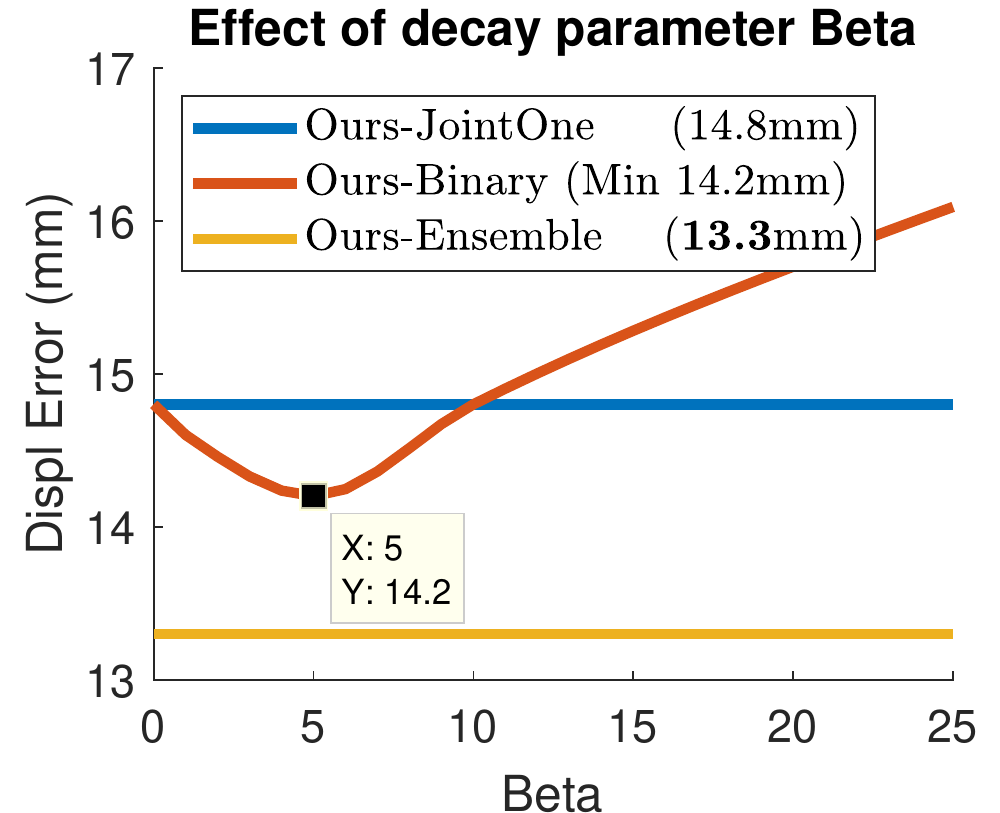}
}{%
  \caption{Performance for the {\it Binary} decay function with respect to $\beta$.}%
  \label{fig.decay_result}
}
\end{floatrow}
\end{figure}

Specifically, the fully connected regression \emph{\textbf{Baseline}} and our attention-based per-pixel regression method are compared in Table~\ref{table.pixel_sampling}. General optical flow (denoted as \emph{\textbf{DenseFlow}}), the \emph{\textbf{SparseFlow}} baseline, and our explicitly learned temporal joint relations (denoted as \emph{\textbf{Ours(Temp)}}) are compared in Table~\ref{table.flow_compare}. Different forms of the pixel weighting function for translation augmentation are investigated in Table~\ref{table.decay_func} and Figure~\ref{fig.decay_result}. We draw conclusions from these experiments one-by-one.

\emph{\textbf{Per-pixel regression with relevant pixel sampling is superior to fully connected regression.}} This can be concluded from Table~\ref{table.pixel_sampling}. Our per-pixel regression using only one pixel located at the prediction of the related joint (denoted as \emph{\textbf{Ours-JointOne}}) outperforms the fully connected \emph{\textbf{Baseline}} by a large margin, $18.4\%$ and $22.1\%$ relative improvement on spatial and temporal joint relations, respectively. Detailed \emph{Bone Error} (left) and \emph{Displ Error} (right) of \emph{Baseline} and \emph{Ours-JointOne} during training are shown in the top row. 

For a more comprehensive understanding, we empirically evaluate several other pixel sampling strategies besides \emph{Ours-JointOne}. \emph{RandOne} randomly samples one pixel; \emph{RandHalf} randomly samples half of the pixels, and \emph{Full} uses all pixels. The results are reported in Table~\ref{table.pixel_sampling}. A couple conclusions can be drawn. \emph{First, sparse connections are superior to holistic mapping.} This can be seen from the lower performance of \emph{Baseline} and \emph{Full} compared to \emph{Ours-JointOne}. \emph{Second, pixel selection matters.} Randomly choosing the sparse connections (\emph{RandOne}, \emph{RandHalf}) yields lower performance than more purposeful selection (\emph{Ours-JointOne}).

\emph{\textbf{Explicitly learned temporal joint relation is superior to optical flow.}} Table~\ref{table.flow_compare} presents a comprehensive comparison between \emph{DenseFlow}, \emph{SparseFlow} and \emph{Ours(Temp)}. \emph{SparseFlow} and \emph{Ours(Temp)} are trained using the same network setting and data for a fair comparison. For \emph{DenseFlow}, we use the state-of-the-art \emph{FlowNet2.0}~\cite{ilg2017flownet} to compute optical flow between two consecutive frames, and then use the same pixel weighting scheme as in \emph{Ours(Temp)}. Since joint displacement in the depth dimension cannot be determined from \emph{DenseFlow}, we directly use the corresponding displacement in depth from \emph{Ours(Temp)} for \emph{DenseFlow}. The \emph{Displ Error} metric is used for evaluation.

Three conclusions can be drawn from Table~\ref{table.flow_compare}. \emph{First, the pixel weighting mechanism that only uses relevant pixels is effective.} This can be concluded from the \emph{SparseFlow} baseline outperforming \emph{DenseFlow} in all the cases, where the relative performance gain is shown as subscripts. \emph{Second, the separate joint relation map for each joint to represent a more explicit joint semantic is effective.} This can be concluded from \emph{Ours(Temp)} further improving \emph{SparseFlow} in all the cases by a large margin, where the relative performance gain to \emph{SparseFlow} is shown as subscripts. \emph{Third, Ours(Temp) is especially superior under large joint displacement.} This can be seen from the larger relative improvement under lower frame rates (\emph{Ours(Temp)} outperforms \emph{SparseFlow} by 33.3\% at 2.5FPS, but by 15.1\% at 25FPS) and for more challenging joints with large displacement (\emph{Ours(Temp)} outperforms \emph{SparseFlow} by 39.4\% in \emph{Hard} level, but by 17.2\% in \emph{Easy} level at 2.5FPS).


\begin{table}
\caption{Comparison to optical flow (\emph{\textbf{DenseFlow}} and \emph{\textbf{SparseFlow}}) for different frame rates. \textbf{Displ Err} is used. \textbf{\emph{Easy}}, \textbf{\emph{Middle}} and \textbf{\emph{Hard}} represent joints whose displacement are less than $30mm$, between $30mm$ and $60mm$, and greater than $60mm$, respectively.}
\begin{center}
\small
\footnotesize
\begin{tabular}{l | l | l | l | l}
\hline
2.5FPS  & Mean & Easy     & Middle   & Hard     \\
DisplErr(mm)& (Max 39.2) & (62.5\%) & (17.8\%) & (19.7\%) \\
\hline
DenseFlow & 26.8 & 9.52 & 29.7 & 79.0 \\
SparseFlow & $22.2_{\downarrow17.2\%}$ & $7.89_{\downarrow17.1\%}$ & $27.3_{\downarrow8.08\%}$ &  $62.7_{\downarrow20.6\%}$ \\
Ours(Temp) & $\textbf{14.8}_{\downarrow33.3\%}$ & $\textbf{6.53}_{\downarrow17.2\%}$ & $\textbf{18.2}_{\downarrow33.3\%}$ & $\textbf{38.0}_{\downarrow39.4\%}$ \\
\hline
\end{tabular}
\begin{tabular}{l | l | l | l | l}
\hline
8FPS  & Mean & Easy & Middle & Hard\\
DisplErr(mm) & (Max 12.7) & (88.9\%) & (8.49\%) & (2.65\%) \\
\hline
DenseFlow & 9.08   & 6.19 & 25.4 & 53.8 \\
SparseFlow & $7.85_{\downarrow13.5\%}$ & $5.36_{\downarrow13.4\%}$ & $22.7_{\downarrow10.6\%}$ & $44.0_{\downarrow18.2\%}$ \\
Ours(Temp) & $\textbf{6.07}_{\downarrow22.7\%}$ & $\textbf{4.39}_{\downarrow13.4\%}$ & $\textbf{15.9}_{\downarrow30.0\%}$ & $\textbf{30.7}_{\downarrow30.2\%}$ \\
\hline
\end{tabular}
\begin{tabular}{l | l | l | l | l}
\hline
25FPS  & Mean & Easy & Middle& Hard\\
DisplErr(mm) & (Max 4.32) & (99.1\%) & (0.85\%) & (0.02\%)  \\
\hline
DenseFlow & 3.44 &3.25 & 24.5 & 48.3 \\
SparseFlow & $3.05_{\downarrow11.3\%}$ & $2.87_{\downarrow11.7\%}$ & $22.3_{\downarrow8.98\%}$ & $46.1_{\downarrow4.55\%}$\\
Ours(Temp) & $\textbf{2.59}_{\downarrow15.1\%}$ & $\textbf{2.46}_{\downarrow14.3\%}$ & $\textbf{17.0}_{\downarrow23.8\%}$ & $\textbf{34.9}_{\downarrow24.3\%}$ \\
\hline
\end{tabular}
\end{center}
\label{table.flow_compare}
\end{table}



\emph{\textbf{Decay function for translation augmentation is effective.}} We investigate four forms of decay function $\mathcal{O}(\mathbf{J}, \beta)$ in Eq.~\ref{eq.weight_map_general}. They are \emph{Binary}, \emph{Gaussian}, \emph{Linear} and \emph{Exponential} defined in Table~\ref{table.decay_func}. When $\beta=0 \text{ or } \infty$, $\mathcal{O}(\mathbf{J}, \beta)$ degenerates to \emph{Ours-JointOne} and \emph{Full}, respectively. It is seen that when $\beta$ is slightly larger than $0$, we obtain better performance ($\beta=5$ in the {\it Binary} case). Similar observations are obtained using other decay functions. The performance differences between different decay functions are minor, but an ensemble of models with different functions yields considerable improvement. We thus use this \emph{\textbf{Ours-Ensemble}} model as our final joint relation estimator for tracking.

\begin{table}
\caption{Effects of spatiotemporal joint relations on tracking. (See supplement for detailed results on all joints and dimensions.)}
\begin{center}
\footnotesize
\begin{tabular}{l | l | l | l }
\hline
\textbf{Joint Error} (mm) & HM36 & HM36MPII & description \\
\hline
Integral Reg~\cite{sun2018integral} & 63.8 & 49.5 & single frame baseline \\
\hline
Ours-F              & 61.7 & 48.8 & +$\mathbb{D}_{f}$\\
Ours-FB             & 61.1 & 48.5 & +$\mathbb{D}_{fb}$\\
Ours-MF             & 59.6 & 47.8 & +$\mathbb{D}_{mf}$\\
\hline
Ours-Temporal       & $59.3_{\downarrow7.1\%}$ & $47.7_{\downarrow3.6\%}$ & +$\mathbb{D}_{mfb}$\\
Ours-Spatial        & $62.1_{\downarrow2.7\%}$ & $47.4_{\downarrow4.2\%}$ & +bone vector\\
\hline
Ours-Spatiotemporal & $\textbf{58.1}_{\downarrow8.9\%}$ & $\textbf{45.5}_{\downarrow8.1\%}$ & +$\mathbb{D}_{mfb}$ +bone vector\\
\hline
\end{tabular}
\end{center}
\label{table.tracking_hm36}
\end{table}



\begin{table*}
\footnotesize
\centering
\begin{tabular}{|l|c|c|c|c|c|c|c|c|c|c|c|c|c|c|}
\hline
Method  &Tome&Moreno&Zhou&Jahangiri&Mehta&Martinez&Kanazawa&Fang&Sun& S{\'a}r{\'a}ndi &Sun&Dabral& Hossain&Ours\\
 &        ~\cite{tome2017lifting} & ~\cite{moreno20163d} & ~\cite{zhou2017monocap}&     ~\cite{jahangiri2017generating}&  ~\cite{mehta2016monocular}&~\cite{martinez2017simple}  &~\cite{kanazawa2017end} &~\cite{fang2018learning} &~\cite{sun2017compositional}&~\cite{sarandi2018synthetic}$^\dagger$&~\cite{sun2018integral}&~\cite{dabral2018learning}$^*$&~\cite{hossain2017exploiting}$^*$  &$^*$\\
\hline
\textbf{MPJPE}  & 88.4 & 87.3& 79.9  & 77.6  & 72.9  & 62.9 & 88.0  & 60.4& 59.1 & 54.2& \emph{49.6} & \underline{52.1}& \underline{51.9}&\textbf{\underline{45.5}}\\
\hline
\end{tabular}
\caption{Comparisons to previous work on Human3.6M. All methods used extra
2D training data. Methods with $^*$ exploit temporal information. Ours gives the lowest error and improves on the state of the art~\cite{sun2018integral} by 4.1mm (relative $8.3\%$).}%
\label{table.hm36_sota}
\end{table*}

\begin{table*}
\footnotesize
\centering
\begin{tabular}{|l|c|c|c|c|c|c|c|c|c|c}
\hline
Method & Zhou~\cite{zhou2016sparseness} &Tekin~\cite{tekin2016direct}&Xingyi~\cite{zhou2016deep}&Sun~\cite{sun2017compositional}&Pavlakos~\cite{pavlakos2016coarse}&Sun~\cite{sun2018integral}&Lin~\cite{lin2017recurrent}$^*$&Coskun~\cite{coskun2017long}$^*$&Ours$^*$\\
\hline
\textbf{MPJPE} &113.0 &125.0  &107.3  & 92.4& 71.9 & \emph{64.1} &\underline{73.1}&\underline{71.0}& \textbf{\underline{58.1}}\\
\hline
\end{tabular}
\caption{Comparison to previous work on Human3.6M. No extra training data is used. Methods with $^*$ exploit temporal information. Ours yields the lowest error and improves on the state of the art~\cite{sun2018integral} by 6.0mm (relative $9.4\%$).}%
\label{table.hm36_only_sota}
\end{table*}

\textbf{Effects of Spatiotemporal Joint Relations on Tracking.} Now we apply the explicitly learned spatiotemporal joint relations to the tracking framework introduced in Section~\ref{sec.tracking} for better joint localization in videos. Hence, the joint localization metric \emph{Joint Error} is used for evaluation. For the single frame pose estimator, we use the Pytorch implementation~\cite{integralgit} of Integral Regression~\cite{sun2018integral}. Specifically, two single-frame baselines are used. The first uses only \emph{HM36} data for training and employs a two-stage network architecture. The second baseline mixes \emph{HM36} and \emph{MPII} data for training and uses a one-stage network architecture (\emph{HM36MPII}). \emph{Note that these two baselines are the state of the art and thus produce strong baseline results.}

Table~\ref{table.tracking_hm36} shows how joint localization accuracy improves using our explicitly learned spatiotemporal joint relations. The \emph{Joint Error} metric and 8FPS frame rates are used for evaluation. We can draw three conclusions. First, multi-step tracking is effective. With more elements in the duration set, better performance is obtained ($\mathbb{D}_{mfb}>\mathbb{D}_{mf}>\mathbb{D}_{fb}>\mathbb{D}_{f}$). In other words, the more observations we get from different durations in Eq.~\ref{eq.err_func_pro}, the better pose tracking result we can get. Second, both \emph{\textbf{Ours-Spatial}} (or \emph{\textbf{Ours-S}}) and \emph{\textbf{Ours-Temporal}} (or \emph{\textbf{Ours-T}}) effectively improve all single-frame baselines. Larger relative improvement can be obtained on the low-performance single-frame baseline. Third, combining both spatial and temporal relations (\emph{\textbf{Ours-Spatiotemporal}} or \emph{\textbf{Ours-ST}}) leads to further improvements and performs the best.

\textbf{Comparison with the state of the art} Previous works are commonly divided into two categories. The first uses extra 2D data for training. The second only uses \emph{HM36} data for training. They are compared to our method in Table~\ref{table.hm36_sota} and Table~\ref{table.hm36_only_sota}, respectively.  Methods marked with * are tracking based methods that exploit temporal information. Note that in Table~\ref{table.hm36_sota}, although S{\'a}r{\'a}ndi et al.~\cite{sarandi2018synthetic}$^\dagger$ do not use extra 2D pose ground truth, they use extra 2D images for synthetic occlusion data augmentation where the occluders are realistic object segments from the Pascal VOC 2012 dataset~\cite{everingham2011pascal}. Their data augmentation technique is also complementary to our method.

Our method sets a new state of the art on the Human3.6M benchmark under both categories. Specifically, it improves the state-of-the-art by 4.1mm (relative 8.3\%) in Table~\ref{table.hm36_sota}, and 6.0mm (relative 9.4\%) in Table~\ref{table.hm36_only_sota}.

\subsection{Experiments on MSRA15}
\label{sec.exp_msra15}

 
\emph{\textbf{MSRA Hand 2015 (MSRA15)~\cite{sun2015cascaded}}} is a standard benchmark for 3D hand pose estimation from depth. It consists of 76.5k video frames in total. For evaluation, we use two common accuracy metrics in the literature~\cite{sun2015cascaded,wan2016hand,chen2017pose,ge20173d,guo2017region,oberweger2017deepprior++,wan2017crossing,wan2018dense}. The first one is mean 3D \emph{Joint Error}. The second one is the percentage of correct frames (\emph{\textbf{PCF}}). A frame is considered correct if its maximum joint error is less than a small threshold~\cite{sun2015cascaded}. For single-frame pose estimation, many recent works provide their single-frame predictions~\cite{wan2018dense,chen2018shpr,ge2018point,ge2018hand}. We use all of them as our single-frame baselines. In addition, we re-implement a new single-frame baseline using Integral Regression~\cite{sun2018integral}.




\begin{figure}
\begin{floatrow}
\capbtabbox[.4\textwidth]
{
\footnotesize
\begin{tabular}{l | l }
\hline
\textbf{Joint Error} (mm) &  IntReg~\cite{sun2018integral}  \\
\hline
single frame & 8.42 \\
\hline
Ours-F              & 7.10 \\
Ours-FB             & 6.78 \\
Ours-MF             & 6.42 \\
\hline
Ours-T       & $6.35_{\downarrow24.6\%}$ \\
Ours-S        & $6.67_{\downarrow20.8\%}$ \\
\hline
Ours-ST & $\textbf{5.96}_{\downarrow29.2\%}$\\
\hline
\end{tabular}
}
{%
}
\ffigbox[.6\textwidth]
{
  \includegraphics [width=0.8\linewidth] {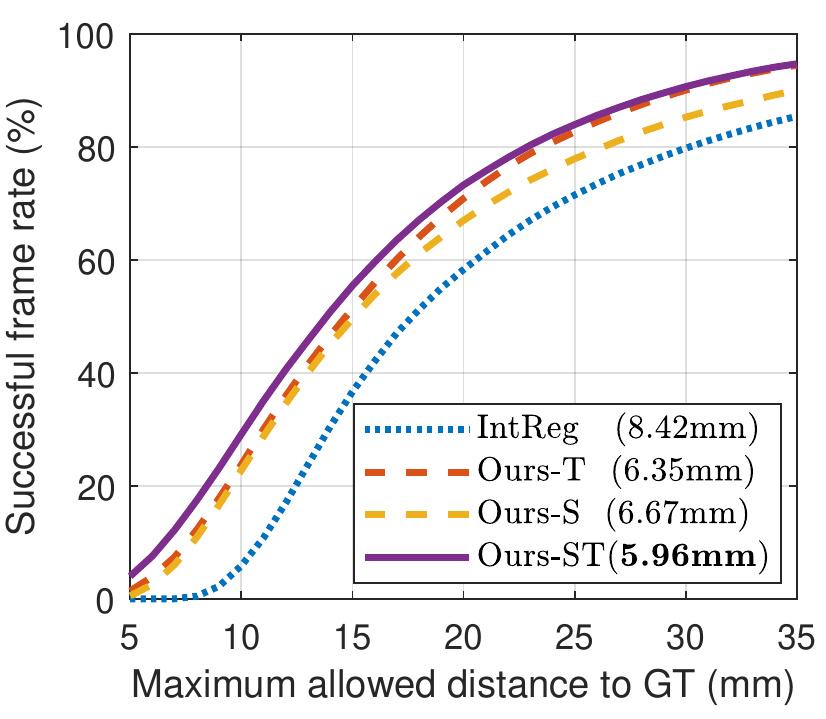}
}
{
}
\end{floatrow}
\capbtabbox[1.0\textwidth]
{
\footnotesize
\quad
\begin{tabular}{l | l | l | l | l }
\hline
\textbf{Joint Err}& PointNet & SHPR-Net & Point2Point & DenseReg \\
(mm) & ~\cite{ge2018hand} & ~\cite{chen2018shpr} & ~\cite{ge2018point} & ~\cite{wan2018dense} \\
\hline
single  & 8.50 & 7.76 & 7.71 & 7.23 \\
\hline
Ours-T  & $7.19_{\downarrow15.4\%}$ & $7.02_{\downarrow9.54\%}$ & $6.80_{\downarrow11.8\%}$ & $6.96_{\downarrow3.73\%}$ \\
Ours-S  & $7.54_{\downarrow11.3\%}$ & $7.09_{\downarrow8.63\%}$ & $6.89_{\downarrow10.6\%}$ & $6.45_{\downarrow10.8\%}$ \\
\hline
Ours-ST & $\textbf{6.68}_{\downarrow21.4\%}$ & $\textbf{6.56}_{\downarrow15.5\%}$& $\textbf{6.28}_{\downarrow18.5\%}$&$\textbf{6.44}_{\downarrow10.9\%}$\\
\hline
\end{tabular}
}
{
  \caption{Effects of spatiotemporal joint relations on tracking. The graph shows a comparison to IntReg~\cite{sun2018integral} on the \emph{PCF} metric.}
  \label{table.tracking_msra15}
}
\end{figure}

\textbf{Effect of Spatiotemporal Joint Relations on Tracking}
We apply our explicitly learned joint relations on all of the single-frame baselines for tracking, and list the results in Table~\ref{table.tracking_msra15}. The mean 3D joint error is used as the evaluation metric. Our observations are similar to or even stronger than those for the \emph{HM36} experiments. 

First, our spatial and/or temporal joint relations significantly improve all single-frame baselines. The relative performance gains over the single-frame baselines are shown in the subscripts. Second, the improvement is especially large on the per-pixel joint classification baseline~\cite{sun2018integral}. Specifically, our method improves the Integral Regression baseline by 2.46mm (relative 29.2\%), establishing a new record of 5.96mm mean 3D joint error on this benchmark. 

\textbf{Comparisons to the state of the art} We compare our method to previous works in Table~\ref{table.msra15_sota} with the mean 3D joint error metric. Our method achieves a \emph{5.96mm} mean 3D joint error and outperforms the previous state-of-the-art result of $7.23mm$~\cite{wan2018dense} by a large margin (17.6\% relative improvement). Please see the supplement for more comparisons using the \emph{PCF} metric.






\begin{table}
\caption{Comparison to previous work on MSRA15. Ours yields the lowest error and improves~\cite{wan2018dense} by 1.27mm (relative $17.6\%$).}
\begin{center}
\footnotesize
\begin{tabular}{l | l | l | l | l }
\hline
Methods & Sun~\cite{sun2015cascaded} &Ge~\cite{ge2016robust}&Madadi~\cite{madadi2017occlusion}&Wan~\cite{wan2017crossing}\\
\textbf{Joint Error}&15.2mm&13.2mm&12.8mm&12.2mm\\
\hline
\hline
Wang~\cite{wang2018region}&Ge~\cite{ge20173d}&Oberweger~\cite{oberweger2017deepprior++}&Chen~\cite{chen2017pose}& Ge~\cite{ge2018hand}\\
9.79mm&9.58mm&9.50mm&8.65mm& 8.50mm \\
\hline
\hline
Chen~\cite{chen2018shpr} & Ge~\cite{ge2018point} &Moon~\cite{moon2018v2v}& Wan ~\cite{wan2018dense}& Ours\\
7.76mm & 7.71mm & 7.49mm & 7.23mm &
\textbf{5.96mm} \\
\hline
\end{tabular}
\end{center}
\label{table.msra15_sota}
\end{table}


{\small
\bibliographystyle{ieee}
\bibliography{egbib}

\begin{thebibliography}{10}\itemsep=-1pt

\bibitem{integralgit}
{Integral Regression Pytorch Github}.
\newblock \url{https://github.com/JimmySuen/integral-human-pose}.

\bibitem{rgb3dhandgit}
{SHP Evaluation Toolkit Github}.
\newblock \url{https://github.com/lmb-freiburg/hand3d}.

\bibitem{bogo2016keep}
F.~Bogo, A.~Kanazawa, C.~Lassner, P.~Gehler, J.~Romero, and M.~J. Black.
\newblock Keep it smpl: Automatic estimation of 3d human pose and shape from a
  single image.
\newblock In {\em European Conference on Computer Vision}, pages 561--578.
  Springer, 2016.

\bibitem{bulat2016human}
A.~Bulat and G.~Tzimiropoulos.
\newblock Human pose estimation via convolutional part heatmap regression.
\newblock In {\em European Conference on Computer Vision}, pages 717--732.
  Springer, 2016.

\bibitem{cai2018weakly}
Y.~Cai, L.~Ge, J.~Cai, and J.~Yuan.
\newblock Weakly-supervised 3d hand pose estimation from monocular rgb images.
\newblock {\em ECCV, Springer}, 12, 2018.

\bibitem{cao2016realtime}
Z.~Cao, T.~Simon, S.-E. Wei, and Y.~Sheikh.
\newblock Realtime multi-person 2d pose estimation using part affinity fields.
\newblock {\em arXiv preprint arXiv:1611.08050}, 2016.

\bibitem{carreira2016human}
J.~Carreira, P.~Agrawal, K.~Fragkiadaki, and J.~Malik.
\newblock Human pose estimation with iterative error feedback.
\newblock In {\em Proceedings of the IEEE Conference on Computer Vision and
  Pattern Recognition}, pages 4733--4742, 2016.

\bibitem{chen20163d}
C.-H. Chen and D.~Ramanan.
\newblock 3d human pose estimation= 2d pose estimation+ matching.
\newblock {\em arXiv preprint arXiv:1612.06524}, 2016.

\bibitem{chen2017pose}
X.~Chen, G.~Wang, H.~Guo, and C.~Zhang.
\newblock Pose guided structured region ensemble network for cascaded hand pose
  estimation.
\newblock {\em arXiv preprint arXiv:1708.03416}, 2017.

\bibitem{chen2018shpr}
X.~Chen, G.~Wang, C.~Zhang, T.-K. Kim, and X.~Ji.
\newblock Shpr-net: Deep semantic hand pose regression from point clouds.
\newblock {\em IEEE Access}, 6:43425--43439, 2018.

\bibitem{chu2016structured}
X.~Chu, W.~Ouyang, H.~Li, and X.~Wang.
\newblock Structured feature learning for pose estimation.
\newblock In {\em Proceedings of the IEEE Conference on Computer Vision and
  Pattern Recognition}, pages 4715--4723, 2016.

\bibitem{coskun2017long}
H.~Coskun, F.~Achilles, R.~S. DiPietro, N.~Navab, and F.~Tombari.
\newblock Long short-term memory kalman filters: Recurrent neural estimators
  for pose regularization.
\newblock In {\em ICCV}, pages 5525--5533, 2017.

\bibitem{dabral2018learning}
R.~Dabral, A.~Mundhada, U.~Kusupati, S.~Afaque, A.~Sharma, and A.~Jain.
\newblock Learning 3d human pose from structure and motion.
\newblock In {\em Proceedings of the European Conference on Computer Vision
  (ECCV)}, pages 668--683, 2018.

\bibitem{doering2018joint}
A.~Doering, U.~Iqbal, and J.~Gall.
\newblock Joint flow: Temporal flow fields for multi person tracking.
\newblock {\em arXiv preprint arXiv:1805.04596}, 2018.

\bibitem{einfalt2018activity}
M.~Einfalt, D.~Zecha, and R.~Lienhart.
\newblock Activity-conditioned continuous human pose estimation for performance
  analysis of athletes using the example of swimming.
\newblock {\em arXiv preprint arXiv:1802.00634}, 2018.

\bibitem{everingham2011pascal}
M.~Everingham and J.~Winn.
\newblock The pascal visual object classes challenge 2012 (voc2012) development
  kit.
\newblock {\em Pattern Analysis, Statistical Modelling and Computational
  Learning, Tech. Rep}, 2011.

\bibitem{fabbri2018learning}
M.~Fabbri, F.~Lanzi, S.~Calderara, A.~Palazzi, R.~Vezzani, and R.~Cucchiara.
\newblock Learning to detect and track visible and occluded body joints in a
  virtual world.
\newblock {\em arXiv preprint arXiv:1803.08319}, 2018.

\bibitem{fang2018learning}
H.-S. Fang, Y.~Xu, W.~Wang, X.~Liu, and S.-C. Zhu.
\newblock Learning pose grammar to encode human body configuration for 3d pose
  estimation.
\newblock In {\em AAAI}, 2018.

\bibitem{felzenszwalb2005pictorial}
P.~F. Felzenszwalb and D.~P. Huttenlocher.
\newblock Pictorial structures for object recognition.
\newblock {\em International Journal of Computer Vision}, 61(1):55--79, 2005.

\bibitem{fieraru2018learning}
M.~Fieraru, A.~Khoreva, L.~Pishchulin, and B.~Schiele.
\newblock Learning to refine human pose estimation.
\newblock In {\em Proceedings of the IEEE Conference on Computer Vision and
  Pattern Recognition Workshops}, pages 205--214, 2018.

\bibitem{ge2018hand}
L.~Ge, Y.~Cai, J.~Weng, and J.~Yuan.
\newblock Hand pointnet: 3d hand pose estimation using point sets.
\newblock In {\em Proceedings of the IEEE Conference on Computer Vision and
  Pattern Recognition}, pages 8417--8426, 2018.

\bibitem{ge2016robust}
L.~Ge, H.~Liang, J.~Yuan, and D.~Thalmann.
\newblock Robust 3d hand pose estimation in single depth images: from
  single-view cnn to multi-view cnns.
\newblock In {\em Proceedings of the IEEE conference on computer vision and
  pattern recognition}, pages 3593--3601, 2016.

\bibitem{ge20173d}
L.~Ge, H.~Liang, J.~Yuan, and D.~Thalmann.
\newblock 3d convolutional neural networks for efficient and robust hand pose
  estimation from single depth images.
\newblock In {\em Proceedings of the IEEE Conference on Computer Vision and
  Pattern Recognition}, volume~1, page~5, 2017.

\bibitem{ge2018point}
L.~Ge, Z.~Ren, and J.~Yuan.
\newblock Point-to-point regression pointnet for 3d hand pose estimation.
\newblock {\em ECCV, Springer}, 1, 2018.

\bibitem{girdhar2018detect}
R.~Girdhar, G.~Gkioxari, L.~Torresani, M.~Paluri, and D.~Tran.
\newblock Detect-and-track: Efficient pose estimation in videos.
\newblock In {\em Proceedings of the IEEE Conference on Computer Vision and
  Pattern Recognition}, pages 350--359, 2018.

\bibitem{gkioxari2016chained}
G.~Gkioxari, A.~Toshev, and N.~Jaitly.
\newblock Chained predictions using convolutional neural networks.
\newblock In {\em European Conference on Computer Vision}, pages 728--743.
  Springer, 2016.

\bibitem{guo2017region}
H.~Guo, G.~Wang, X.~Chen, C.~Zhang, F.~Qiao, and H.~Yang.
\newblock Region ensemble network: Improving convolutional network for hand
  pose estimation.
\newblock In {\em Image Processing (ICIP), 2017 IEEE International Conference
  on}, pages 4512--4516. IEEE, 2017.

\bibitem{he2017mask}
K.~He, G.~Gkioxari, P.~Dollar, and R.~Girshick.
\newblock Mask r-cnn.
\newblock In {\em International Conference on Computer Vision}, 2017.

\bibitem{he2016deep}
K.~He, X.~Zhang, S.~Ren, and J.~Sun.
\newblock Deep residual learning for image recognition.
\newblock In {\em Proceedings of the IEEE Conference on Computer Vision and
  Pattern Recognition}, pages 770--778, 2016.

\bibitem{hossain2017exploiting}
M.~R.~I. Hossain and J.~J. Little.
\newblock Exploiting temporal information for 3d pose estimation.
\newblock {\em arXiv preprint arXiv:1711.08585}, 2017.

\bibitem{ilg2017flownet}
E.~Ilg, N.~Mayer, T.~Saikia, M.~Keuper, A.~Dosovitskiy, and T.~Brox.
\newblock Flownet 2.0: Evolution of optical flow estimation with deep networks.
\newblock In {\em IEEE conference on computer vision and pattern recognition
  (CVPR)}, volume~2, page~6, 2017.

\bibitem{insafutdinov2017arttrack}
E.~Insafutdinov, M.~Andriluka, L.~Pishchulin, S.~Tang, E.~Levinkov, B.~Andres,
  and B.~Schiele.
\newblock Arttrack: Articulated multi-person tracking in the wild.
\newblock In {\em IEEE Conference on Computer Vision and Pattern Recognition
  (CVPR)}, volume 4327. IEEE, 2017.

\bibitem{insafutdinov2016deepercut}
E.~Insafutdinov, L.~Pishchulin, B.~Andres, M.~Andriluka, and B.~Schiele.
\newblock Deepercut: A deeper, stronger, and faster multi-person pose
  estimation model.
\newblock In {\em European Conference on Computer Vision}, pages 34--50.
  Springer, 2016.

\bibitem{ioffe2015batch}
S.~Ioffe and C.~Szegedy.
\newblock Batch normalization: Accelerating deep network training by reducing
  internal covariate shift.
\newblock {\em arXiv preprint arXiv:1502.03167}, 2015.

\bibitem{ionescu2014human3}
C.~Ionescu, D.~Papava, V.~Olaru, and C.~Sminchisescu.
\newblock Human3. 6m: Large scale datasets and predictive methods for 3d human
  sensing in natural environments.
\newblock {\em IEEE transactions on pattern analysis and machine intelligence},
  36(7):1325--1339, 2014.

\bibitem{iqbal2016posetrack}
U.~Iqbal, A.~Milan, and J.~Gall.
\newblock Posetrack: Joint multi-person pose estimation and tracking.
\newblock {\em arXiv preprint arXiv:1611.07727}, 2016.

\bibitem{iqbal2018hand}
U.~Iqbal, P.~Molchanov, T.~Breuel, J.~Gall, and J.~Kautz.
\newblock Hand pose estimation via latent 2.5 d heatmap regression.
\newblock {\em arXiv preprint arXiv:1804.09534}, 2018.

\bibitem{jahangiri2017generating}
E.~Jahangiri and A.~L. Yuille.
\newblock Generating multiple hypotheses for human 3d pose consistent with 2d
  joint detections.
\newblock {\em arXiv preprint arXiv:1702.02258}, 2017.

\bibitem{jain2014modeep}
A.~Jain, J.~Tompson, Y.~LeCun, and C.~Bregler.
\newblock Modeep: A deep learning framework using motion features for human
  pose estimation.
\newblock In {\em Asian conference on computer vision}, pages 302--315.
  Springer, 2014.

\bibitem{kanazawa2017end}
A.~Kanazawa, M.~J. Black, D.~W. Jacobs, and J.~Malik.
\newblock End-to-end recovery of human shape and pose.
\newblock {\em arXiv preprint arXiv:1712.06584}, 2017.

\bibitem{lin2017recurrent}
M.~Lin, L.~Lin, X.~Liang, K.~Wang, and H.~Cheng.
\newblock Recurrent 3d pose sequence machines.
\newblock In {\em Computer Vision and Pattern Recognition (CVPR), 2017 IEEE
  Conference on}, pages 5543--5552. IEEE, 2017.

\bibitem{luo2017lstm}
Y.~Luo, J.~Ren, Z.~Wang, W.~Sun, J.~Pan, J.~Liu, J.~Pang, and L.~Lin.
\newblock Lstm pose machines.
\newblock {\em arXiv preprint arXiv:1712.06316}, 2017.

\bibitem{madadi2017occlusion}
M.~Madadi, S.~Escalera, A.~Carruesco, C.~Andujar, X.~Bar{\'o}, and
  J.~Gonz{\`a}lez.
\newblock Occlusion aware hand pose recovery from sequences of depth images.
\newblock In {\em Automatic Face \& Gesture Recognition (FG 2017), 2017 12th
  IEEE International Conference on}, pages 230--237. IEEE, 2017.

\bibitem{martinez2017simple}
J.~Martinez, R.~Hossain, J.~Romero, and J.~J. Little.
\newblock A simple yet effective baseline for 3d human pose estimation.
\newblock {\em arXiv preprint arXiv:1705.03098}, 2017.

\bibitem{mehta2016monocular}
D.~Mehta, H.~Rhodin, D.~Casas, O.~Sotnychenko, W.~Xu, and C.~Theobalt.
\newblock Monocular 3d human pose estimation in the wild using improved cnn
  supervision.
\newblock {\em arXiv preprint arXiv:1611.09813}, 2016.

\bibitem{mehta2017vnect}
D.~Mehta, S.~Sridhar, O.~Sotnychenko, H.~Rhodin, M.~Shafiei, H.-P. Seidel,
  W.~Xu, D.~Casas, and C.~Theobalt.
\newblock Vnect: Real-time 3d human pose estimation with a single rgb camera.
\newblock {\em ACM Transactions on Graphics (TOG)}, 36(4):44, 2017.

\bibitem{moon2018v2v}
G.~Moon, J.~Y. Chang, and K.~M. Lee.
\newblock V2v-posenet: Voxel-to-voxel prediction network for accurate 3d hand
  and human pose estimation from a single depth map.
\newblock In {\em CVPR}, volume~2, 2018.

\bibitem{moreno20163d}
F.~Moreno-Noguer.
\newblock 3d human pose estimation from a single image via distance matrix
  regression.
\newblock {\em arXiv preprint arXiv:1611.09010}, 2016.

\bibitem{mueller2018ganerated}
F.~Mueller, F.~Bernard, O.~Sotnychenko, D.~Mehta, S.~Sridhar, D.~Casas, and
  C.~Theobalt.
\newblock Ganerated hands for real-time 3d hand tracking from monocular rgb.
\newblock In {\em Proceedings of the IEEE Conference on Computer Vision and
  Pattern Recognition}, pages 49--59, 2018.

\bibitem{newell2017associative}
A.~Newell, Z.~Huang, and J.~Deng.
\newblock Associative embedding: End-to-end learning for joint detection and
  grouping.
\newblock In {\em Advances in Neural Information Processing Systems}, pages
  2277--2287, 2017.

\bibitem{newell2016stacked}
A.~Newell, K.~Yang, and J.~Deng.
\newblock Stacked hourglass networks for human pose estimation.
\newblock In {\em European Conference on Computer Vision}, pages 483--499.
  Springer, 2016.

\bibitem{oberweger2017deepprior++}
M.~Oberweger and V.~Lepetit.
\newblock Deepprior++: Improving fast and accurate 3d hand pose estimation.
\newblock In {\em ICCV workshop}, volume 840, page~2, 2017.

\bibitem{oikonomidis2011efficient}
I.~Oikonomidis, N.~Kyriazis, and A.~A. Argyros.
\newblock Efficient model-based 3d tracking of hand articulations using kinect.
\newblock In {\em BmVC}, volume~1, page~3, 2011.

\bibitem{ouyang2013joint}
W.~Ouyang and X.~Wang.
\newblock Joint deep learning for pedestrian detection.
\newblock In {\em Proceedings of the IEEE International Conference on Computer
  Vision}, pages 2056--2063, 2013.

\bibitem{panteleris2018using}
P.~Panteleris, I.~Oikonomidis, and A.~Argyros.
\newblock Using a single rgb frame for real time 3d hand pose estimation in the
  wild.
\newblock In {\em 2018 IEEE Winter Conference on Applications of Computer
  Vision (WACV)}, pages 436--445. IEEE, 2018.

\bibitem{papandreou2017towards}
G.~Papandreou, T.~Zhu, N.~Kanazawa, A.~Toshev, J.~Tompson, C.~Bregler, and
  K.~Murphy.
\newblock Towards accurate multi-person pose estimation in the wild.
\newblock {\em arXiv preprint arXiv:1701.01779}, 2017.

\bibitem{paszke2017automatic}
A.~Paszke, S.~Gross, S.~Chintala, G.~Chanan, E.~Yang, Z.~DeVito, Z.~Lin,
  A.~Desmaison, L.~Antiga, and A.~Lerer.
\newblock Automatic differentiation in pytorch.
\newblock 2017.

\bibitem{pavlakos2016coarse}
G.~Pavlakos, X.~Zhou, K.~G. Derpanis, and K.~Daniilidis.
\newblock Coarse-to-fine volumetric prediction for single-image 3d human pose.
\newblock {\em arXiv preprint arXiv:1611.07828}, 2016.

\bibitem{pfister2015flowing}
T.~Pfister, J.~Charles, and A.~Zisserman.
\newblock Flowing convnets for human pose estimation in videos.
\newblock In {\em Proceedings of the IEEE International Conference on Computer
  Vision}, pages 1913--1921, 2015.

\bibitem{pishchulin2016deepcut}
L.~Pishchulin, E.~Insafutdinov, S.~Tang, B.~Andres, M.~Andriluka, P.~V. Gehler,
  and B.~Schiele.
\newblock Deepcut: Joint subset partition and labeling for multi person pose
  estimation.
\newblock In {\em Proceedings of the IEEE Conference on Computer Vision and
  Pattern Recognition}, pages 4929--4937, 2016.

\bibitem{qian2014realtime}
C.~Qian, X.~Sun, Y.~Wei, X.~Tang, and J.~Sun.
\newblock Realtime and robust hand tracking from depth.
\newblock In {\em Proceedings of the IEEE conference on computer vision and
  pattern recognition}, pages 1106--1113, 2014.

\bibitem{rogez2016mocap}
G.~Rogez and C.~Schmid.
\newblock Mocap-guided data augmentation for 3d pose estimation in the wild.
\newblock In {\em Advances in Neural Information Processing Systems}, pages
  3108--3116, 2016.

\bibitem{sarandi2018synthetic}
I.~S{\'a}r{\'a}ndi, T.~Linder, K.~O. Arras, and B.~Leibe.
\newblock Synthetic occlusion augmentation with volumetric heatmaps for the
  2018 eccv posetrack challenge on 3d human pose estimation.
\newblock {\em arXiv preprint arXiv:1809.04987}, 2018.

\bibitem{song2017thin}
J.~Song, L.~Wang, L.~Van~Gool, and O.~Hilliges.
\newblock Thin-slicing network: A deep structured model for pose estimation in
  videos.
\newblock In {\em The IEEE Conference on Computer Vision and Pattern
  Recognition (CVPR)}, volume~2, 2017.

\bibitem{spurr2018cross}
A.~Spurr, J.~Song, S.~Park, and O.~Hilliges.
\newblock Cross-modal deep variational hand pose estimation.
\newblock In {\em Proceedings of the IEEE Conference on Computer Vision and
  Pattern Recognition}, pages 89--98, 2018.

\bibitem{sridhar2016real}
S.~Sridhar, F.~Mueller, M.~Zollh{\"o}fer, D.~Casas, A.~Oulasvirta, and
  C.~Theobalt.
\newblock Real-time joint tracking of a hand manipulating an object from rgb-d
  input.
\newblock In {\em European Conference on Computer Vision}, pages 294--310.
  Springer, 2016.

\bibitem{sun2017compositional}
X.~Sun, J.~Shang, S.~Liang, and Y.~Wei.
\newblock Compositional human pose regression.
\newblock In {\em International Conference on Computer Vision}, 2017.

\bibitem{sun2015cascaded}
X.~Sun, Y.~Wei, S.~Liang, X.~Tang, and J.~Sun.
\newblock Cascaded hand pose regression.
\newblock In {\em Proceedings of the IEEE conference on computer vision and
  pattern recognition}, pages 824--832, 2015.

\bibitem{sun2018integral}
X.~Sun, B.~Xiao, F.~Wei, S.~Liang, and Y.~Wei.
\newblock Integral human pose regression.
\newblock In {\em Proceedings of the European Conference on Computer Vision
  (ECCV)}, pages 529--545, 2018.

\bibitem{tekin2016structured}
B.~Tekin, I.~Katircioglu, M.~Salzmann, V.~Lepetit, and P.~Fua.
\newblock Structured prediction of 3d human pose with deep neural networks.
\newblock {\em arXiv preprint arXiv:1605.05180}, 2016.

\bibitem{tekin2016direct}
B.~Tekin, A.~Rozantsev, V.~Lepetit, and P.~Fua.
\newblock Direct prediction of 3d body poses from motion compensated sequences.
\newblock In {\em Proceedings of the IEEE Conference on Computer Vision and
  Pattern Recognition}, pages 991--1000, 2016.

\bibitem{tome2017lifting}
D.~Tome, C.~Russell, and L.~Agapito.
\newblock Lifting from the deep: Convolutional 3d pose estimation from a single
  image.
\newblock {\em arXiv preprint arXiv:1701.00295}, 2017.

\bibitem{tompson2014joint}
J.~J. Tompson, A.~Jain, Y.~LeCun, and C.~Bregler.
\newblock Joint training of a convolutional network and a graphical model for
  human pose estimation.
\newblock In {\em Advances in neural information processing systems}, pages
  1799--1807, 2014.

\bibitem{vaswani2017attention}
A.~Vaswani, N.~Shazeer, N.~Parmar, J.~Uszkoreit, L.~Jones, A.~N. Gomez,
  {\L}.~Kaiser, and I.~Polosukhin.
\newblock Attention is all you need.
\newblock In {\em Advances in Neural Information Processing Systems}, pages
  5998--6008, 2017.

\bibitem{wan2017crossing}
C.~Wan, T.~Probst, L.~Van~Gool, and A.~Yao.
\newblock Crossing nets: Combining gans and vaes with a shared latent space for
  hand pose estimation.
\newblock In {\em 2017 IEEE Conference on Computer Vision and Pattern
  Recognition (CVPR)}. IEEE, 2017.

\bibitem{wan2018dense}
C.~Wan, T.~Probst, L.~Van~Gool, and A.~Yao.
\newblock Dense 3d regression for hand pose estimation.
\newblock In {\em Proc. IEEE Conf. Comput. Vis. Pattern Recognit.(CVPR)}, pages
  1--10, 2018.

\bibitem{wan2016hand}
C.~Wan, A.~Yao, and L.~Van~Gool.
\newblock Hand pose estimation from local surface normals.
\newblock In {\em European conference on computer vision}, pages 554--569.
  Springer, 2016.

\bibitem{wang2018region}
G.~Wang, X.~Chen, H.~Guo, and C.~Zhang.
\newblock Region ensemble network: Towards good practices for deep 3d hand pose
  estimation.
\newblock {\em Journal of Visual Communication and Image Representation}, 2018.

\bibitem{wei2016convolutional}
S.-E. Wei, V.~Ramakrishna, T.~Kanade, and Y.~Sheikh.
\newblock Convolutional pose machines.
\newblock In {\em Proceedings of the IEEE Conference on Computer Vision and
  Pattern Recognition}, pages 4724--4732, 2016.

\bibitem{xiao2018simple}
B.~Xiao, H.~Wu, and Y.~Wei.
\newblock Simple baselines for human pose estimation and tracking.
\newblock {\em arXiv preprint arXiv:1804.06208}, 2018.

\bibitem{xiu2018pose}
Y.~Xiu, J.~Li, H.~Wang, Y.~Fang, and C.~Lu.
\newblock Pose flow: Efficient online pose tracking.
\newblock {\em arXiv preprint arXiv:1802.00977}, 2018.

\bibitem{yang2016end}
W.~Yang, W.~Ouyang, H.~Li, and X.~Wang.
\newblock End-to-end learning of deformable mixture of parts and deep
  convolutional neural networks for human pose estimation.
\newblock In {\em CVPR}, 2016.

\bibitem{yasin2016dual}
H.~Yasin, U.~Iqbal, B.~Kruger, A.~Weber, and J.~Gall.
\newblock A dual-source approach for 3d pose estimation from a single image.
\newblock In {\em Proceedings of the IEEE Conference on Computer Vision and
  Pattern Recognition}, pages 4948--4956, 2016.

\bibitem{zhang20163d}
J.~Zhang, J.~Jiao, M.~Chen, L.~Qu, X.~Xu, and Q.~Yang.
\newblock 3d hand pose tracking and estimation using stereo matching.
\newblock {\em arXiv preprint arXiv:1610.07214}, 2016.

\bibitem{zhou2017towards}
X.~Zhou, Q.~Huang, X.~Sun, X.~Xue, and Y.~Wei.
\newblock Towards 3d human pose estimation in the wild: a weakly-supervised
  approach.
\newblock In {\em International Conference on Computer Vision}, 2017.

\bibitem{zhou2016deep}
X.~Zhou, X.~Sun, W.~Zhang, S.~Liang, and Y.~Wei.
\newblock Deep kinematic pose regression.
\newblock In {\em Computer Vision--ECCV 2016 Workshops}, pages 186--201.
  Springer, 2016.

\bibitem{zhou2016model}
X.~Zhou, Q.~Wan, W.~Zhang, X.~Xue, and Y.~Wei.
\newblock Model-based deep hand pose estimation.
\newblock {\em arXiv preprint arXiv:1606.06854}, 2016.

\bibitem{zhou2016sparseness}
X.~Zhou, M.~Zhu, S.~Leonardos, K.~G. Derpanis, and K.~Daniilidis.
\newblock Sparseness meets deepness: 3d human pose estimation from monocular
  video.
\newblock In {\em Proceedings of the IEEE Conference on Computer Vision and
  Pattern Recognition}, pages 4966--4975, 2016.

\bibitem{zhou2017monocap}
X.~Zhou, M.~Zhu, G.~Pavlakos, S.~Leonardos, K.~G. Derpanis, and K.~Daniilidis.
\newblock Monocap: Monocular human motion capture using a cnn coupled with a
  geometric prior.
\newblock {\em arXiv preprint arXiv:1701.02354}, 2017.

\bibitem{zimmermann2017learning}
C.~Zimmermann and T.~Brox.
\newblock Learning to estimate 3d hand pose from single rgb images.
\newblock In {\em International Conference on Computer Vision}, volume~1,
  page~3, 2017.

\end{thebibliography}
}

\pagebreak
\begin{center}
\textbf{\large Supplementary Materials: Explicit Spatiotemporal Joint Relation Learning for Tracking Human Pose}
\end{center}
\setcounter{equation}{0}
\setcounter{figure}{0}
\setcounter{table}{0}
\setcounter{section}{0}
\setcounter{page}{0}


\section{Additional Results}
\label{sec.more_exp}

Due to page limitations, some experimental results were mentioned but not included in the main paper, and are reported in this supplement instead.

\subsection{Experiments on Human3.6M}

\begin{table*} 
\centering
\begin{tabular}{l||c|c||c|c||c|c}
\hline
FPS &  \multicolumn{2}{|c}{2.5}&  \multicolumn{2}{|c}{8}&  \multicolumn{2}{|c}{25}\\
\hline
Method & DenseFlow& Ours(Temp)& DenseFlow& Ours(Temp) & DenseFlow& Ours(Temp) \\
\hline \hline
Average    	    & 26.8 & $14.8_{\downarrow44.8\%}$  & 9.08  & $6.07_{\downarrow33.1\%}$& 3.44  & $2.59_{\downarrow24.7\%}$\\
\hline
Ankle    & 41.0 & $18.3_{\downarrow22.7mm}$ & 14.0 & $8.78_{\downarrow5.22mm}$ & 5.23 & $4.01_{\downarrow1.22mm}$\\
Knee     & 26.0 & $13.1_{\downarrow12.9mm}$ & 9.14 & $5.83_{\downarrow3.31mm}$ & 3.42 & $2.54_{\downarrow0.88mm}$\\
Hip 	 & 9.67 & $5.42_{\downarrow4.25mm}$ & 3.37 & $2.17_{\downarrow1.20mm}$ & 1.17 & $0.89_{\downarrow0.28mm}$\\
Torso    & 12.8 & $8.57_{\downarrow4.23mm}$ & 4.91 & $3.71_{\downarrow1.20mm}$ & 1.81 & $1.60_{\downarrow0.21mm}$\\
Neck     & 16.9 & $11.3_{\downarrow5.60mm}$ & 6.24 & $4.70_{\downarrow1.54mm}$ & 2.47 & $2.07_{\downarrow0.40mm}$\\
Head     & 21.1 & $13.1_{\downarrow8.00mm}$ & 7.75 & $5.37_{\downarrow2.38mm}$ & 3.39 & $2.44_{\downarrow0.95mm}$\\
Wrist    & 54.7 & $31.7_{\downarrow23.0mm}$ & 16.8 & $11.7_{\downarrow5.10mm}$ & 6.17 & $4.57_{\downarrow1.60mm}$\\
Elbow    & 40.1 & $21.6_{\downarrow18.5mm}$ & 13.2 & $8.27_{\downarrow4.93mm}$ & 4.93 & $3.45_{\downarrow1.48mm}$\\
Shoulder & 21.0 & $13.0_{\downarrow8.00mm}$ & 7.57 & $5.36_{\downarrow2.21mm}$ & 3.00 & $2.40_{\downarrow0.60mm}$\\
\hline
x        & 16.0 & $5.78_{\downarrow10.2mm}$ & 5.00 & $2.36_{\downarrow2.64mm}$ & 1.81 & $1.03_{\downarrow0.78mm}$\\
y        & 12.3 & $5.41_{\downarrow6.89mm}$ & 3.92 & $2.15_{\downarrow1.77mm}$ & 1.41 & $0.92_{\downarrow0.49mm}$\\
z        & - & 10.3 & - & 4.25 & - & 1.80 \\
\hline
\end{tabular}
\caption{Comparison to optical flow (\emph{DenseFlow}) for different frame rates on all joints and dimensions. The relative performance gain in $mm$ is shown in the subscript. Extension of Table 3 (main paper).}
\label{table.joint_error_dmpjpe}
\end{table*}

\begin{table*} 
\centering
\begin{tabular}{l||c|c||c|c}
\hline
Method & HM36~\cite{sun2018integral}& Ours-Temporal & HM36+MPII~\cite{sun2018integral}& Ours-Temporal\\
\hline \hline
Average    	    & 63.8  & $59.3_{\downarrow7.1\%}$& 49.5  & $47.7_{\downarrow3.6\%}$\\
\hline
Ankle    & 82.4  & $76.6_{\downarrow5.8mm}$  & 65.7 & $64.0_{\downarrow1.7mm}$\\
Knee     & 55.0  & $50.6_{\downarrow4.4mm}$  & 43.0 & $41.1_{\downarrow1.9mm}$\\
Hip 	 & 28.4  & $24.6_{\downarrow3.8mm}$  & 22.2 & $20.3_{\downarrow1.9mm}$\\
Torso    & 44.6  & $42.9_{\downarrow1.7mm}$  & 38.3 & $37.4_{\downarrow0.9mm}$\\
Neck     & 68.4  & $65.5_{\downarrow2.9mm}$  & 54.5 & $53.5_{\downarrow1.0mm}$\\
Head     & 69.8  & $66.2_{\downarrow3.6mm}$  & 48.9 & $47.6_{\downarrow1.3mm}$\\
Wrist    & 100.1 & $91.4_{\downarrow8.7mm}$  & 73.8 & $69.4_{\downarrow4.4mm}$\\
Elbow    & 80.4  & $74.8_{\downarrow5.6mm}$  & 61.8 & $59.7_{\downarrow2.1mm}$\\
Shoulder & 68.9  & $65.0_{\downarrow3.9mm}$  & 55.4 & $54.2_{\downarrow1.2mm}$\\
\hline
x        & 19.8  & $18.5_{\downarrow1.3mm}$  & 13.8 & $13.4_{\downarrow0.4mm}$\\
y        & 17.4  & $16.2_{\downarrow1.2mm}$  & 14.3 & $13.8_{\downarrow0.5mm}$\\
z        & 50.5  & $47.3_{\downarrow3.2mm}$  & 40.2 & $38.8_{\downarrow1.4mm}$\\
\hline
\end{tabular}
\caption{Detailed results on all joints and dimensions for single-frame baselines and Ours-Temporal methods. The relative performance gain in $mm$ is shown in the subscript. Extension of Table 5 (main paper).}
\label{table.joint_error_mpjpe}
\end{table*}

\paragraph{Detailed Results on All Joints and Dimensions} In Table 3 of the main paper, we show that \emph{Ours(Temp)} outperforms \emph{DenseFlow} under all frame rates using the same pixel weighting strategy, with especially large differences under large joint displacement. Table~\ref{table.joint_error_dmpjpe} further reports the performance improvements from \emph{Ours(Temp)} to \emph{DenseFlow} on all the joints and dimensions. The performance gain in $mm$ is shown as subscripts. Since we directly use the corresponding deformation in depth from \emph{Ours(Temp)} for \emph{DenseFlow}, comparisons in the depth dimension are not reported. The conclusions are consistent with Table 3 in the main paper. First, \emph{Ours(Temp)} clearly outperforms \emph{DenseFlow} under all frame rates for all the joints and dimensions. Second, \emph{Ours(Temp)} is especially superior to \emph{DenseFlow} under large joint displacement. This can be seen from the larger relative improvements under lower frame rates and those end joints with large displacements, e.g., Ankle, Head and Wrist. Additionally, it is seen that improvement in the x dimension is larger than in the y dimension. This is due to more frequent body movements in the x dimension than in the y dimension in the dataset.

In Table 5 of the main paper, we show that \emph{Ours-Temporal} effectively and consistently improves all single-frame baselines, namely \emph{HM36} and \emph{HM36+MPII}. Table~\ref{table.joint_error_mpjpe} further reports the performance improvement of \emph{Ours-Temporal} on these single-frame baselines for all the joints and dimensions. We can conclude that our method effectively improves the accuracy for all the joints and dimensions, especially for challenging ones like wrist, elbow, ankle joint and z dimension.

\begin{figure*}
\includegraphics [width=0.6\linewidth] {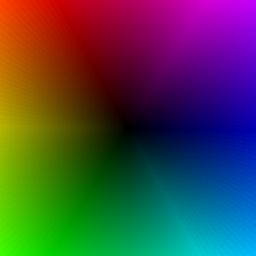}
\caption{(Best viewed in color) Color map for optical flow and joint displacement visualization.}
\label{fig.color_map}
\end{figure*}

\begin{figure*}
\includegraphics [width=1.0\linewidth] {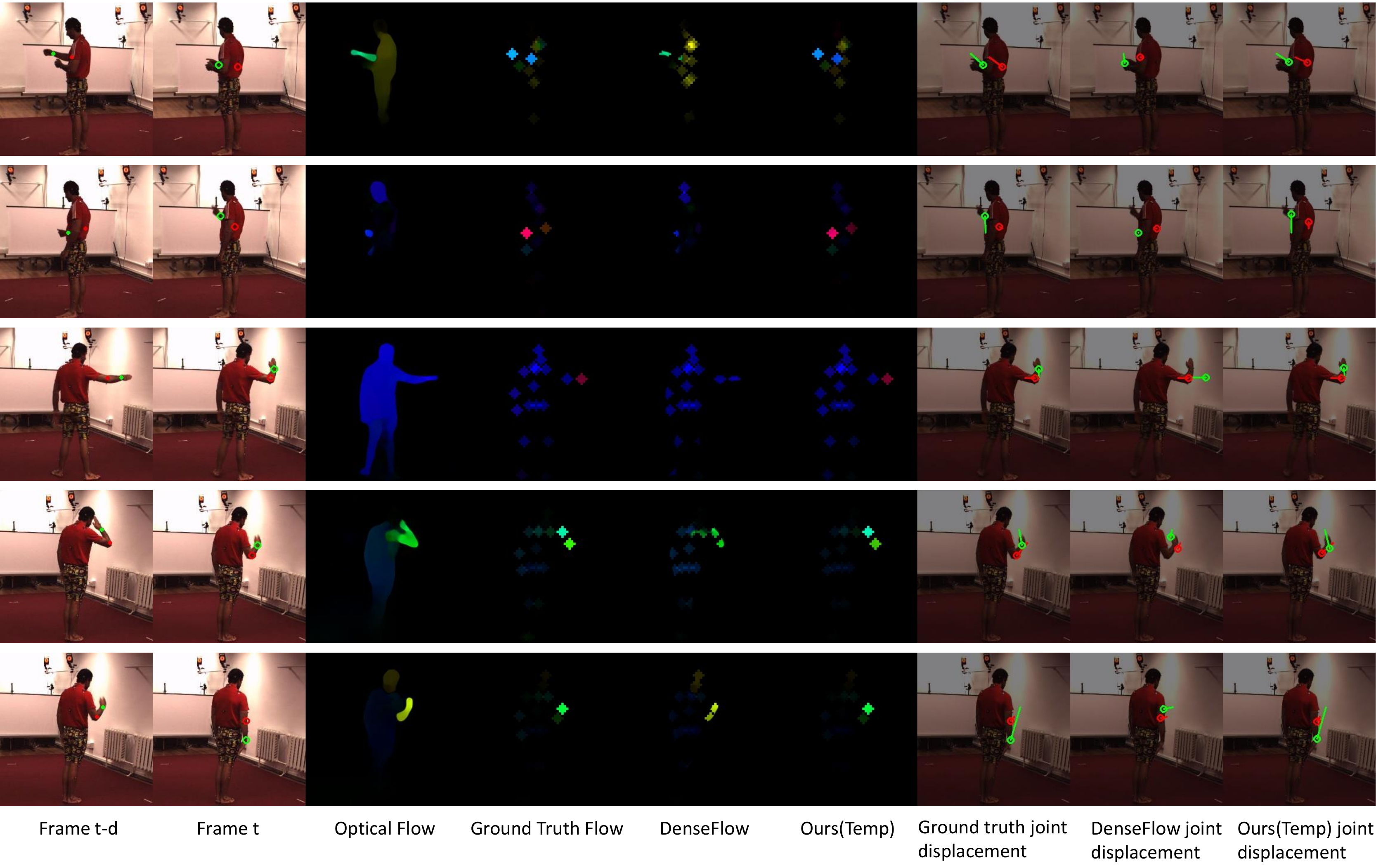}
\caption{(Best viewed in color) Qualitative comparison results of \emph{Ours(Temp)} and optical flow (\emph{DenseFlow}). Samples from subject 9, action Directions, camera view 1. Displacements of Right Wrist and Right Elbow joints are shown by green and red arrows, respectively. See our supplementary demo video for more vivid and dynamic results.}
\label{fig.qualitative_result_1}
\end{figure*}

\begin{figure*}
\includegraphics [width=1.0\linewidth] {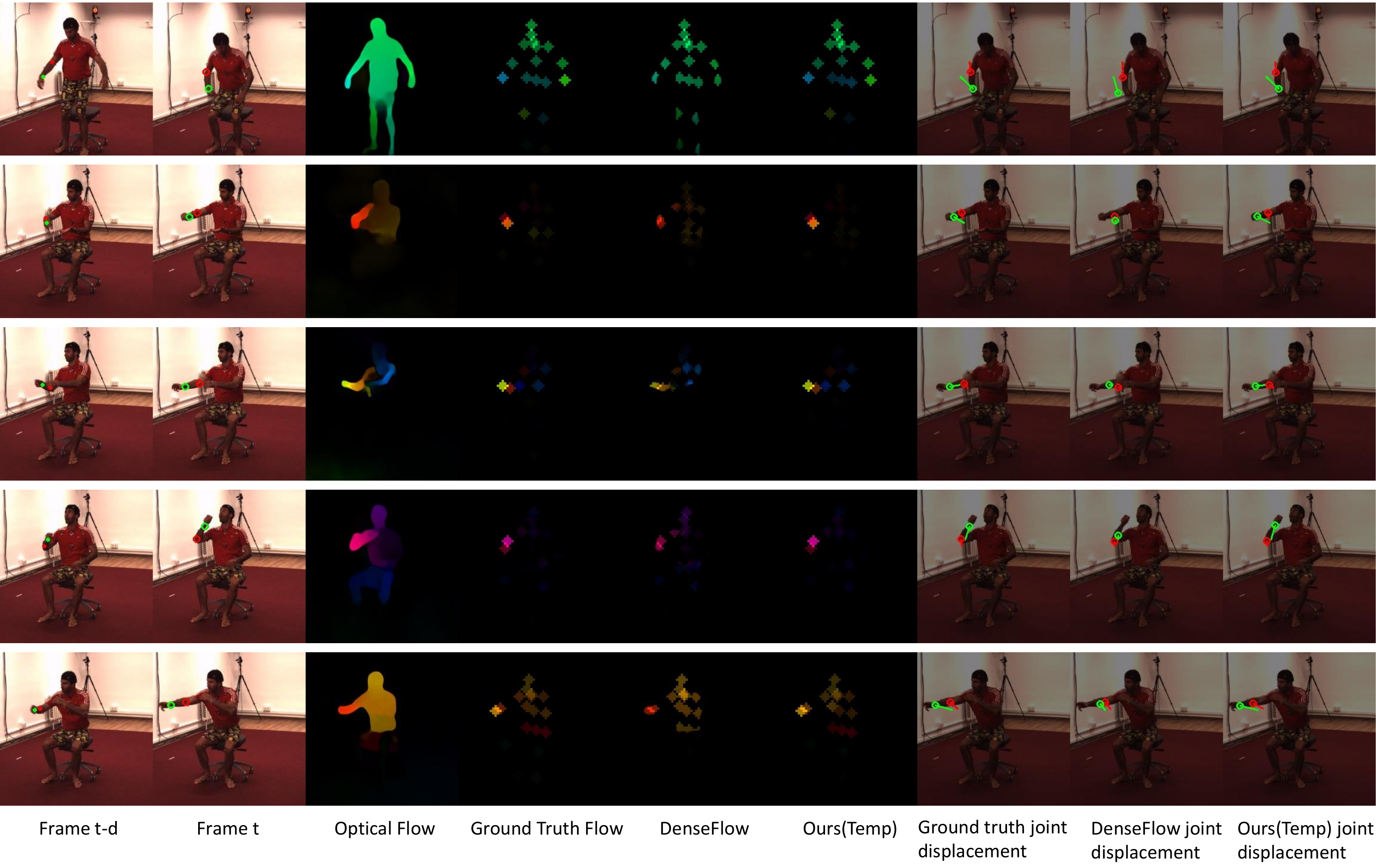}
\caption{(Best viewed in color) Qualitative comparison results of \emph{Ours(Temp)} and optical flow (\emph{DenseFlow}). Samples from subject 9, action Eating, camera view 2. Displacements of Right Wrist and Right Elbow joints are shown by green and red arrows, respectively. See our supplementary demo video for more vivid and dynamic results.}
\label{fig.qualitative_result_2}
\end{figure*}

\begin{figure*}
\includegraphics [width=1.0\linewidth] {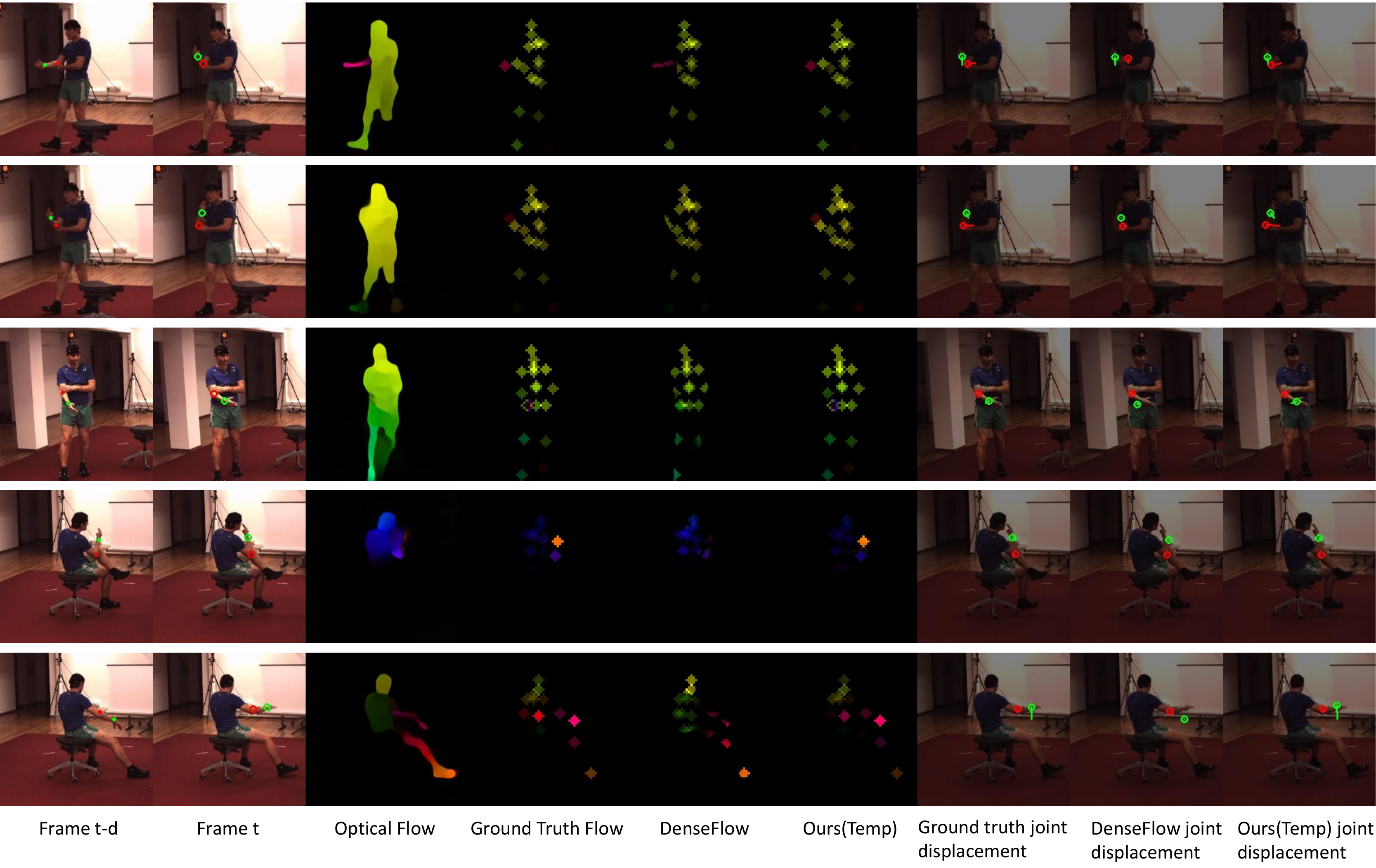}
\caption{(Best viewed in color) Qualitative comparison results of \emph{Ours(Temp)} and optical flow (\emph{DenseFlow}). Samples from subject 11, action Smoking, camera view 3. Displacements of Right Wrist and Right Elbow joints are shown by green and red arrows, respectively. See our supplementary demo video for more vivid and dynamic results.}
\label{fig.qualitative_result_3}
\end{figure*}

\begin{figure*}
\includegraphics [width=1.0\linewidth] {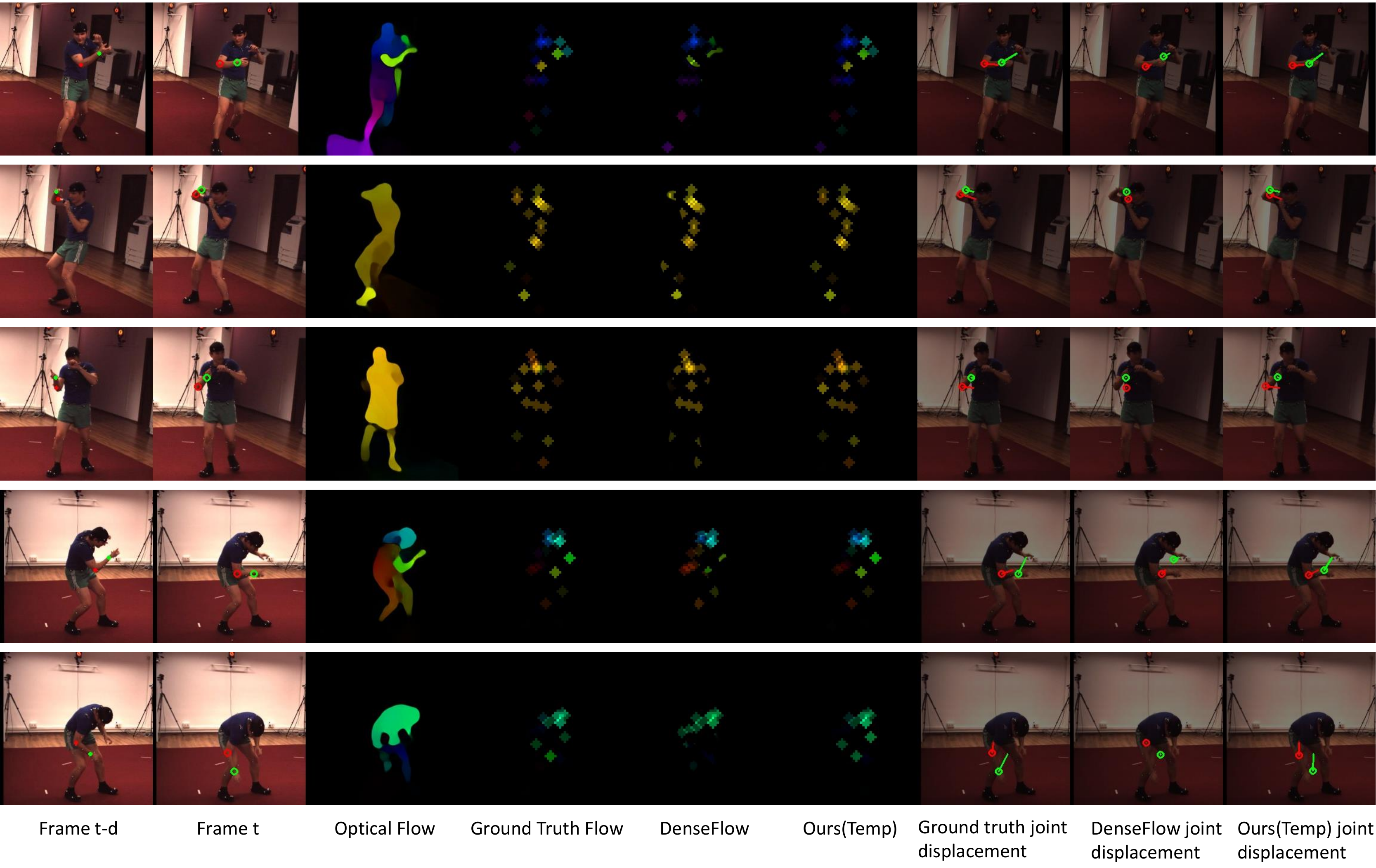}
\caption{(Best viewed in color) Qualitative comparison results of \emph{Ours(Temp)} and optical flow (\emph{DenseFlow}). Samples from subject 11, action Taking photo, camera view 4. Displacements of Right Wrist and Right Elbow joints are shown by green and red arrows, respectively. See our supplementary demo video for more vivid and dynamic results.}
\label{fig.qualitative_result_4}
\end{figure*}

\paragraph{Qualitative Results} Figures~\ref{fig.qualitative_result_1}, ~\ref{fig.qualitative_result_2}, ~\ref{fig.qualitative_result_3} and ~\ref{fig.qualitative_result_4} show additional qualitative comparison results of \emph{Ours-Temporal} and \emph{DenseFlow}. Displacements of Right Wrist and Right Elbow joints are shown by green and red arrows, respectively. Figure~\ref{fig.color_map} is the color map for optical flow and joint displacement visualization. See our supplementary demo video for more vivid and dynamic results.

\subsection{Experiments on MSRA15}
\begin{figure}
\includegraphics [width=1.0\linewidth] {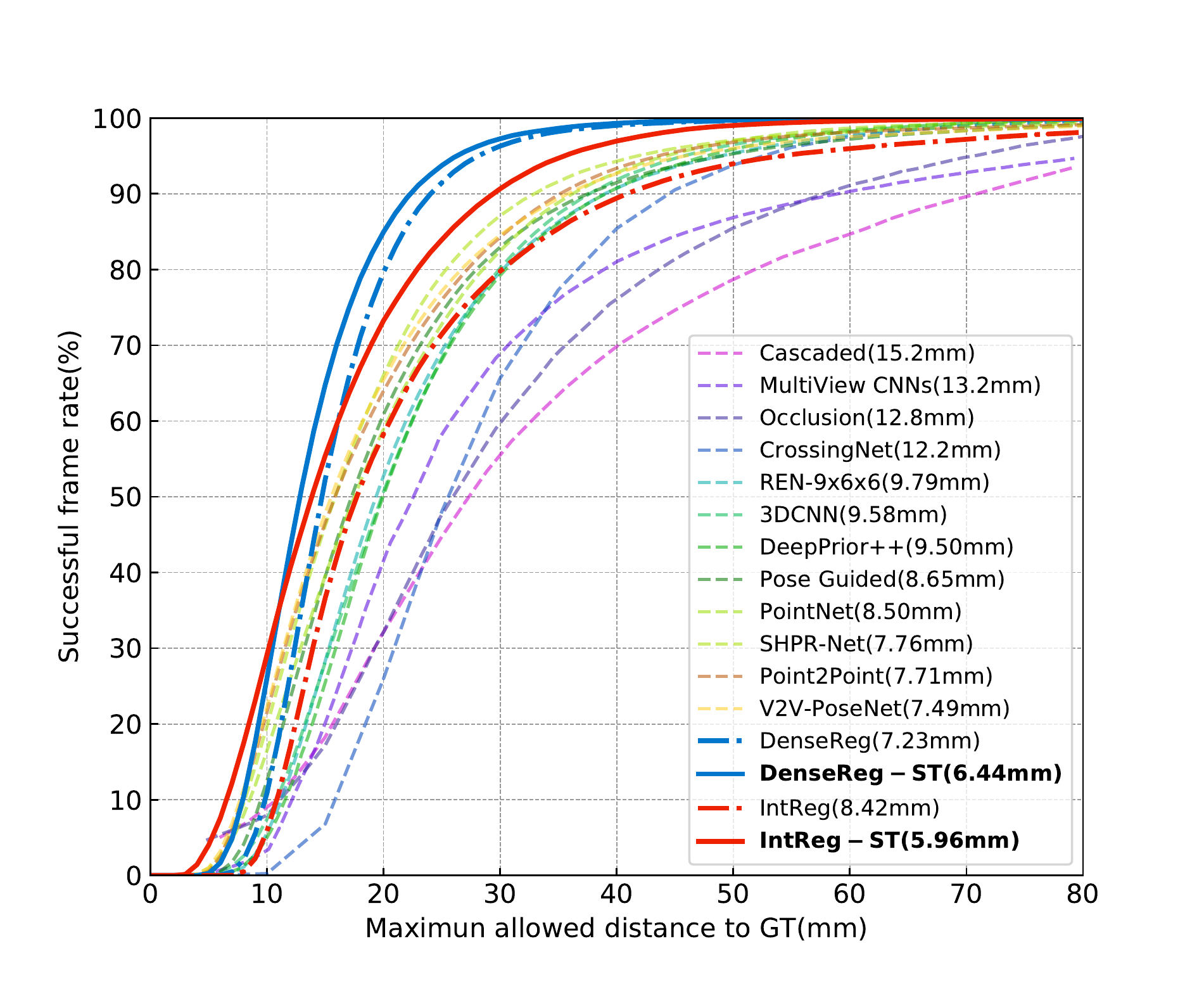}
\caption{Comparisons to the state of the art on MSRA15 under the metric of \emph{PCF}.}
\label{fig.msra15_sota}
\end{figure}

\paragraph{Comparisons to the state of the art} In Table 9 of the main paper, we compare our method to previous works with the mean 3D joint error metric. Figure~\ref{fig.msra15_sota} further shows the performance improvement of \emph{Ours-ST} on IntReg~\cite{sun2018integral} (\emph{IntReg-ST}) and DenseReg~\cite{wan2018dense} (\emph{DenseReg-ST}) under the metric of \emph{PCF}. Our method sets a new state of the art on the MSRA15 benchmark.

\subsection{Experiments on SHP and D+O}

\begin{figure}
\begin{floatrow}
\capbtabbox[.4\textwidth]
{
  \footnotesize
    \begin{tabular}{lll}
    \hline
    Method     & AUC            & \begin{tabular}[c]{@{}c@{}}EPE\\ mean\end{tabular}         \\ \hline
    Panteleris~\cite{panteleris2018using}  & 0.941      & -              \\
    Z\&B ~\cite{zimmermann2017learning}            & 0.948      & 8.68             \\
    Mueller~\cite{mueller2018ganerated}    & 0.965      & -            \\
    Spurr~\cite{spurr2018cross}            & 0.983      & 8.56         \\
    Iqbal~\cite{iqbal2018hand}             & 0.994      & -           \\
    Cai~\cite{cai2018weakly}               & 0.994      & -           \\
    \hline
    Ours single                              & 0.994     & 7.82     \\
    Ours track                                  & \textbf{0.995}      & \textbf{7.62}  \\ \hline
    \end{tabular}
}
{
  \caption{Comparsions to state-of-the-art on SHP.}
  \label{table.shpresult}
}
\capbtabbox[.6\textwidth]
{
    \footnotesize
    \begin{tabular}{lccc}
    \hline
    Method         & AUC  & \multicolumn{2}{c}{\begin{tabular}[c]{@{}c@{}}EPE\\ mean  median\end{tabular}} \\ \hline
    Mueller~\cite{mueller2018ganerated} & 0.56 &      -                          & -                       \\
    Iqbal~\cite{iqbal2018hand}  & 0.71 & 31.9                           & 25.4                                          \\
    \hline
    Ours single  & 0.72 &  28.9                      &    24.7     \\          
    Ours track      & \textbf{0.73} & \textbf{27.5}   & \textbf{24.3}\\ 
    \hline
    \end{tabular}
}
{%
  \caption{Comparisons to state-of-the-art on D+O.}%
  \label{table.doresult}
}
\end{floatrow}
\end{figure}

\emph{\textbf{Dexter+Object~\cite{sridhar2016real} (D+O)}} and \emph{\textbf{Stereo Hand Pose~\cite{zhang20163d} (SHP)}} are two standard 3D hand pose datasets composed of RGB-D images. \emph{D+O} provides six test video sequences recorded using a static camera with a single hand interacting with an object. \emph{SHP} provides 3D annotations of 21 keypoints for 18k stereo frame pairs, recording a person performing various gestures in front of six different backgrounds. \emph{Note that only the RGB images are used in our experiments.} 

We follow the evaluation protocol and implementation provided by~\cite{zimmermann2017learning} that uses 15k samples from the \emph{SHP} dataset for training (\emph{SHP-train}) and the remaining 3k samples for evaluation (\emph{SHP-eval}). The model trained using \emph{SHP-train} is also evaluated on \emph{D+O}. For evaluation metrics, we use average End-Point-Error (EPE) and the Area Under Curve (AUC) on the Percentage of Correct Keypoints (PCK) as done in~\cite{iqbal2018hand,mueller2018ganerated,zimmermann2017learning,rgb3dhandgit}. For single-frame pose estimation, we re-implement a new baseline using Integral Regression~\cite{sun2018integral}. 


\textbf{Comparison to the state of the art} We compare our method to the state of the art on the \emph{SHP} and \emph{D+O} datasets in Table~\ref{table.shpresult} and Table~\ref{table.doresult}, respectively. AUC and EPE are used as evaluation metrics. On both benchmarks, our re-implemented single-frame pose estimator already achieves state of the art performance, but then our tracking method brings consistent improvements to the single-frame baselines, setting the new state-of-the-art in performance. 

\section{Implementation Details}
\label{sec.imp_detail}

\subsection{Training Details on Different Datasets}
\textbf{\emph{Human3.6M}} For RGB image based human pose tracking on the \emph{HM36} dataset, ResNet~\cite{he2016deep} (ResNet-50 by default) is adopted as the backbone network as in~\cite{sun2018integral}. The head network for the joint displacement map is fully convolutional. It first uses deconvolution layers ($4\times4$ kernels, stride 2) to upsample the feature map to the required resolution ($64\times64$ by default). The number of output channels is fixed to 256 as in~\cite{he2017mask,sun2018integral}. Then, a $1\times1$ convolutional layer is used to produce $K$ joint displacement maps. PyTorch~\cite{paszke2017automatic} is used for implementation. Adam is employed for optimization. The input image is normalized to $256\times256$ by default. Data augmentation includes random translation ($\pm2\%$ of the image size), scale ($\pm25\%$), rotation ($\pm30$ degrees) and flip. In all experiments, the base learning rate is 1e-5. It drops to 1e-6 when the loss on the validation set saturates. Four GPUs are utilized. The mini-batch size is 128, and batch normalization~\cite{ioffe2015batch} is used. 

\textbf{\emph{MSRA Hand 2015}} For depth image based hand pose tracking on the \emph{MSRA15} dataset, we use the original frame rate with no downsampling. ResNet-18 is adopted as the backbone network, and the input images are normalized to $128\times128$. Data augmentation includes random rotation ($\pm180$ degrees) and scale ($\pm10$\%). No flip and translation augmentation is used. Other training details are the same as in \emph{HM36}.

\emph{\textbf{Dexter+Object}} and \emph{\textbf{Stereo Hand Pose}} For RGB image based hand pose tracking on the \emph{D+O} and \emph{SHP} datasets, the training details including frame rate, backbone network, input image size and augmentation strategies are the same as in \emph{MSRA15}.

\subsection{Ablation Study for Network Feature Sharing}

\begin{figure*}
\ffigbox[1.0\textwidth]
{
\includegraphics [width=1.0\linewidth] {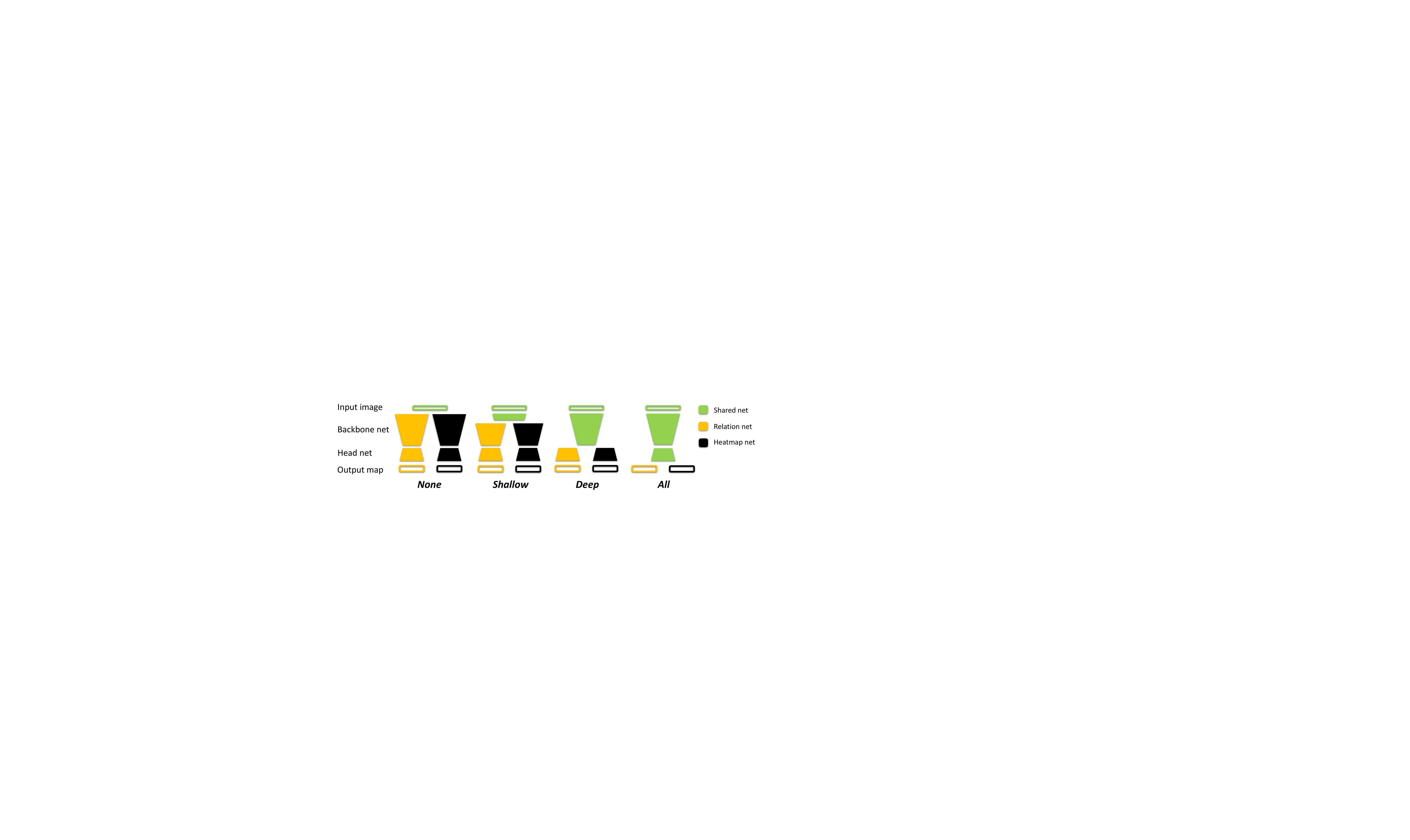}
}
{%
  \caption{Different feature sharing schemes for joint classification and spatial relation regression networks.}
  \label{fig.feature_share}
}
\qquad
\capbtabbox[1.0\textwidth]
{
\begin{tabular}{l | l | l | l | l }
\hline
share methods & None & Shallow &  Deep & All \\
\hline
Bone Error   & 40.4mm & \textbf{39.6}mm & 44.9mm & 108.2mm \\
\hline
Params &  64.9M $>$  &   63.8M $>$   &   42.7M $>$   &  32.7M    \\
\hline
FLOPs  & 10.2G $>$     &   9.01G $>$   &   6.43G $>$   &  6.22G    \\
\hline
\end{tabular}
}{%
  \caption{Performance and computational costs of different feature sharing schemes for joint classification and spatial relation regression networks.}%
  \label{table.feature_share}
}
\end{figure*}

Many previous works~\cite{pishchulin2016deepcut,insafutdinov2016deepercut,papandreou2017towards,fieraru2018learning} propose to learn an additional local offset from a high-score heatmap location to the ground truth location of a joint for self-refinement. The additional offset map usually shares all of the CNN features with the heatmap. Our spatial joint relation network can share CNN features with heatmaps as well. However, it is designed to share only shallow features instead of all of them. This is because shallow features generally represent low-level image elements like colors and edges, while deep features model high-level semantic information specific to the given task. Hence, for joint classification and spatial relation regression, shallow features are shared for computational efficiency, but deep features are separately learned for their particular tasks. We empirically investigate different feature sharing schemes in Figure~\ref{fig.feature_share} and Table~\ref{table.feature_share}.

For implementation, we first train a joint classification (heatmap) network and fix its network weights. Then we add a branch network to the joint classification network at different network depths for spatial relation regression. The later the network branches, the deeper the features that are shared. 

We can conclude from Table~\ref{table.feature_share} that \emph{shallow network feature sharing between heatmap and spatial joint relation is effective.} First, sharing all features and branching at the last 1x1 convolutional layer (denoted as \emph{All}) performs the worst (108.2mm \emph{Bone Error}). This is because it forces the network to perform a linear combination of deep features and map them into both per-pixel joint type and bone vector labels, simultaneously, which ignores the high-nonlinearity between the two target domains. Second, shallow feature sharing with heatmap benefits the learning of spatial joint relations. The \emph{Shallow} architecture (sharing conv1 and conv2 features of Resnet~\cite{he2016deep}) is slightly better than the \emph{None} (no feature is shared) baseline, from 40.4mm to 39.8mm ($1.49\%$ relative improvement) while saving $1.69\%$ parameters and $11.7\%$ FLOPs. Third, when the sharing of features goes deeper, for example to the backbone output (denoted as \emph{Deep}), computational costs are reduced but performance is also decreased. This is because deeper features will have stronger semantics with regard to specific tasks.

\end{document}